\documentclass[12pt]{article}

\usepackage{latexsym}

\long\def\nop#1{}

\expandafter\ifx\csname shortcite\endcsname\relax
\let\shortcite=\cite
\fi

\newbox\current

\long\def\plframebox#1{
\setbox\current\vbox{#1}		

\vbox to \ht\current {\hrule\vss
\hbox to \wd\current {%
\vrule \hss\box\current\hss \vrule}
\vss\hrule }
}

\long\def\eatpar#1{%
\ifx#1\par                      
\let\nextmove=\eatpar           
\else
\let\nextmove=#1
\fi
\noexpand\nextmove
}

\def\modifymargins#1#2{
\newdimen\addtoh
\newdimen\addtow
\addtoh=#1
\addtow=#2

\advance\topmargin by -\addtoh
\multiply\addtoh by 2
\advance\textheight by \addtoh

\advance\oddsidemargin by -\addtow
\advance\evensidemargin by -\addtow
\multiply\addtow by 2
\advance\textwidth by \addtow
}

\begingroup
\catcode`\~=11
\gdef\centertilde#1{\lower #1pt\hbox{~}}
\endgroup

\newcount\currenttime
\newcount\hour
\newcount\minute

\def\printtime{%
\currenttime=\time
\hour=\currenttime
\divide\hour by 60
\minute=-\hour
\multiply\minute by 60
\advance\minute by \currenttime
\the\hour:\ifnum\minute<10 0\fi\the\minute
}

\begingroup
\makeatletter
\global\let\@@date=\@date
\gdef\@date{\@@date\ --- \printtime}
\endgroup

\def\oggi{\number\day\space 
\ifcase\month\or
Gennaio\or Febbraio\or Marzo\or Aprile\or Maggio\or Giugno\or
Luglio\or Agosto\or Settembre\or Ottobre\or Novembre\or Dicembre\fi
\space \number\year}

\newcounter{rmexample}

\def\proof{\noindent {\sl Proof.\ \ }}

\def\qed{\hfill{\boxit{}}
  \ifdim\lastskip<\medskipamount \removelastskip\penalty55\medskip\fi}
\def\qedn#1{\hfill{\boxit{}$_#1$}
  \ifdim\lastskip<\medskipamount \removelastskip\penalty55\medskip\fi}
\long\def\boxit#1{\vbox{\hrule\hbox{\vrule\kern3pt
                  \vbox{\kern3pt#1\kern3pt}\kern3pt\vrule}\hrule}}

\def\ie{i.e.}
\def\eg{e.g.}
\def\wrt{w.r.t.}

\def\l{\langle}
\def\r{\rangle}

\def\true{{\sf true}}
\def\false{{\sf false}}

\def\np{{\rm NP}}
\def\conp{{\rm coNP}}
\def\Dp{${\rm D}^p$}
\def\S#1{\mbox{$\Sigma^p_{#1}$}}
\def\P#1{$\Pi^p_{#1}$}

\def\Dlog#1{\mbox{$\Delta^p_{#1}[\log n]$}}

\def\profont{\sf}

\def\sat{{\profont sat}}
\def\unsat{{\profont unsat}}

\def\x3c{{\profont x3c}}

\def\possnewtheorem#1#2{
\expandafter\ifx\csname #1\endcsname\relax
\newtheorem{#1}{#2}
\fi
}

\def\possnewtheoremthree#1[#2]#3{
\expandafter\ifx\csname #1\endcsname\relax
\newtheorem{#1}[#2]{#3}
\fi
}

\possnewtheorem{theorem}{Theorem}
\possnewtheorem{corollary}{Corollary}
\possnewtheorem{lemma}{Lemma}
\possnewtheoremthree{proposition}[theorem]{Proposition}
\possnewtheorem{definition}{Definition}
\possnewtheorem{question}{Question}
\possnewtheorem{example}{Example}
\possnewtheorem{property}{Property}
\possnewtheorem{assumption}{Assumption}
\possnewtheorem{conjecture}{Conjecture}
\newenvironment{theorem*}[1]{{\noindent \bf Theorem~#1}\begin{it}}{\end{it}\

}

\def\qequiv#1{\stackrel{#1}{\equiv}}

\long\def\ies#1{I.E.S.%
\if.#1%
\else
#1%
\fi}

\long\def\comment#1\endcomment{\vskip\baselineskip
{\tt

#1
}\vskip\baselineskip}

\long\def\comment#1\endcomment{}

\title{Redundancy in Logic I:\\
CNF Propositional Formulae}
\author{Paolo Liberatore%
\thanks{
Dipartimento di Informatica e Sistemistica,
Universit\`a di Roma ``La Sapienza'',
via Salaria 113, 00198, Rome, Italy.}
}

\begin{document}

\maketitle

\begin{abstract}

A knowledge base is redundant if it contains parts that
can be inferred from the rest of it. We study the problem
of checking whether a CNF formula (a set of clauses) is
redundant, that is, it contains clauses that can be derived
from the other ones. Any CNF formula can be made irredundant
by deleting some of its clauses: what
results is an  irredundant equivalent subset (I.E.S.) We
study the complexity of some related problems: verification,
checking existence of a I.E.S.\  with a given size, checking
necessary and possible presence of clauses in I.E.S.'s, and
uniqueness. We also consider the problem of redundancy
with different definitions of equivalence.

\end{abstract}

\


\

\noindent {\bf Note:} this paper is an extended and
revised version of the paper ``The Complexity of
Checking Redundancy of CNF Propositional Formulae'',
presented at the conference ECAI 2002.

\begin{verbatim}
\end{verbatim}

\tableofcontents

\section{Introduction}

A knowledge base is redundant if it contains parts that
can be removed without reducing the information it carries.
In this paper, we study the redundancy of a propositional
formula in Conjunctive Normal Form (CNF), that is, sets
of clauses. A CNF formula is redundant if and only if one
or more clauses can be removed from it without changing
its set of models.

The problem of redundancy, and the related problem of
minimization, are important for a number of reasons.
First, removing redundant clauses leads to a simplification
of the knowledge base. This may have some computational
advantage in some cases (\eg, it leads to an
exponential reduction of size.) Moreover, simplifying
a formula leads to a representation of the same knowledge
that is easier to understand, as a large amount of
redundancy may obscure the meaning of the represented
knowledge. The irredundant part of a knowledge base can
instead be considered the core of the knowledge it
represents.

Redundancy can be a negative characteristic or
not, depending on how the knowledge base is obtained.
Intuitively, a concept that is repeated many times (for
example, in a book) is likely to be a very important
one. If a formula results from the translation of
something expressed by human beings, the fact that a
clause is redundant is noteworthy, as it may
indicate that this clause carries a piece of knowledge
that is considered important.

On the other hand, redundancy may be a negative feature
of a knowledge base, as it may result from an incorrect
encoding or merging of several sources. In such cases,
indeed, it is possible that the intended meaning of a
clause is different from what the clause formally means
(for example, the clause has been expressed using the wrong
variable names.) Whatever the reason a clause is redundant,
the fact that it is redundant is an hint of something,
which may be either an high importance of the knowledge
it express, or an hint of a mistake that has been made
while building the knowledge base.

The problem of redundancy of knowledge bases may also be
relevant to applications in which efficiency of entailment is
important. Indeed, the size of a knowledge base is one
of the factors that determine the speed of the inference
process. While some theorem provers introduce a limited
number of redundant formulae for speeding up solving,
excessive redundancy can cause problems of storage,
which in turns slows down reasoning. In particular, updates
can increase the size of knowledge bases exponentially
\cite{cado-etal-99,libe-00}, and redundancy makes the problem of
storing the knowledge base worst.

Algorithms for checking
redundancy of knowledge bases have been developed for the
case of production rules \cite{gins-88-a,schm-snyd-97}.
In this paper, we instead study redundancy of propositional
knowledge base in CNF form, that is, checking whether a
clause in a set is implied by the others.

A related question that has been already investigated
in the propositional case is whether a knowledge base is
equivalent to a shorter one. This problem is called
{\em minimization of propositional formulae}, and it has been
one of the first to be analyzed from the point of view
of computational complexity: its study begun in the paper that
introduced the polynomial hierarchy \cite{meye-stoc-72}.
A complexity characterization of this problem has been first
given for Horn knowledge bases
\cite{maie-80,ausi-etal-86,hamm-koga-93}; afterwards, the
problem has been tackled again in the general case
\cite{hema-wech-97,uman-97}. While the Horn case is
now quite understood (the problem is \np-complete,
using several different notions of minimality,) some
problems regarding non-Horn formulae are still open.
For example, the problem of deciding whether a formula
is minimal (no other formula with less literals is
equivalent to it) is trivially in \S{2}, but
has only be proved \conp-hard quite recently
\cite{hema-wech-97}, and no other strict bound is
known. What makes this problem difficult to handle
is the fact that the considered formulae are not
constrained to any particular form, such as CNF or
DNF, or even NNF.

Redundancy elimination can be considered as a weak form
of formula minimization: if a set of clauses is redundant,
it is not minimal, as some clauses can be removed from
it while preserving equivalence. On the other hand,
redundancy elimination only allows for removal of clauses,
so it is not guaranteed to produce a minimal knowledge base.
For example, $\{ x \vee y, x \vee \neg y\}$ is irredundant,
but is equivalent to a shorter set: $\{x\}$. A related
problem, not analyzed in this paper, is that of removing redundancy
from a single clause, that is, removing literals from clauses
rather than removing clauses from sets. The computational
analysis of this problem, and of related ones, has been
done by Gottlob and Ferm\"uller \shortcite{gott-ferm-93}.

The problem of redundancy elimination is relevant for at
least two reasons. First, it seems somehow easier to remove
redundant clauses, rather than reshaping the whole knowledge
base. Indeed, removing redundant clauses can be done by
checking whether each clause can be inferred by the other
ones, while finding a minimal equivalent formula
involves a process of {\em guessing and checking} a whole
knowledge base for equivalence. Even for short knowledge bases,
the number of candidate equivalent knowledge bases is very high.

A second reason for preferring redundancy elimination to
minimization is that the syntactic form in which
a knowledge base is expressed can be important. For example,
some semantics for knowledge base revision depend on the
syntax of knowledge bases. If a knowledge base is replaced with
an equivalent one, even a single update can lead to a completely
different result \cite{gins-86,nebe-91}.

\begin{table}
\begin{center}
\begin{tabular}{ll}
\hfil Problem &			\hfil Complexity \\
\hline
Checking irredundancy &		\np\  complete \\
A set is an \ies &		\Dp\  complete \\
Existence of an \ies\  of size $\leq k$ &		\S{2}\  complete \\
A clause is in all \ies's &	\np\  complete \\
A clause is in an \ies &	\S{2}\  complete \\
Uniqueness of \ies's &			\Dlog{2}\  complete \\
\hline
\end{tabular}
\caption{Complexity results about redundancy}
\label{complexity-table}
\end{center}
\end{table}

Several problems are related to that of redundancy. The
aim of checking redundancy is to end up with a subset of
clauses that is both equivalent to the original one
and irredundant. We call it an {\em irredundant equivalent
subset} of the original set, or \ies\  Note that
an \ies\  is a subset of the original set, and can therefore
only contain clauses of the original set.
This makes it different to a minimal equivalent {\em set},
which can instead be composed of arbitrary clauses.

The problems that are analyzed in this paper are: checking
whether a set is an \ies; checking the existence of an \ies\
of size bounded by an integer $k$; deciding whether a clause
is in some, or all, the \ies's; and checking uniqueness.
Table~\ref{complexity-table} contains the complexity of
these problems.

Since redundancy is defined in terms of equivalence (a
formula is redundant if it is equivalent to a proper
subset of its,) alternative definitions of equivalence
lead to different definitions of redundancy. We have
considered two definitions of equivalence, both based
on the sets of entailed formulae. Namely, var-equivalence
\cite{cado-etal-97-c,lang-etal-03} leads to an increase
of complexity, while conditional equivalence
\cite{libe-zhao-03} does not.

\section{Redundancy and \ies's}

In this paper, we study the redundancy
of sets of propositional clauses. A knowledge base
is redundant if it contains some redundant parts, that
is, it is equivalent to one of its proper subsets.
The definition therefore is affected by three factors:

\begin{enumerate}

\item the logic we consider;

\item what is ``a part'' of a knowledge base;

\item the definition of equivalence.

\end{enumerate}

In this paper, we use propositional logic. Nevertheless,
even in this simple case, we still have the problem of
defining what is a part of a knowledge base. For example,
we can consider a knowledge base a set of formulae, and
a part is simply one formula. A restricted case is that
of CNF: a knowledge base is a set of clauses, and a part
is simply a clause. We could also consider generic Boolean
formulae, and a part of them is any subformula.

We however only consider CNF formulae in this paper.
We initially consider the usual definition
of equivalence: other definitions are
considered in a later section. We sometimes use formulae
like
$a_1 \wedge \cdots \wedge a_m \rightarrow b_1 \vee \cdots \vee b_k$,
which can be easily translated into the equivalent clauses
$\neg a_1 \vee \cdots \vee \neg a_m \vee b_1 \vee \cdots \vee b_k$.
We also assume that clauses are not tautological.

\begin{definition}

A CNF formula is a set of non-tautological clauses.

\end{definition}

Clearly, tautologies can be easily checked and removed,
and do not change the complexity of the problems considered
here. The redundancy of a single clause is defined as follows.

\begin{definition}

A clause $\gamma \in \Pi$ is redundant in $\Pi$ if
and only if $\Pi \backslash \{\gamma\} \models \gamma$.

\end{definition}

The redundancy of a clause implies that the clause
can be removed from the set without changing its meaning.
In turns, the redundancy of a set of clauses can be
defined as its equivalence to one of its proper
subsets.

\begin{definition}

A set of clauses $\Pi$ is redundant if and only if
there exists $\Pi' \subset \Pi$ such that
$\Pi' \equiv \Pi$.

\end{definition}

In propositional logic, this definition is equivalent
to the following ones (proofs are omitted due to 
their triviality):

\begin{enumerate}

\item there exists $\Pi' \subset \Pi$ such that
$\Pi' \models \Pi$;

\item $\Pi$ contains a redundant clause.

\end{enumerate}

These definitions are equivalent in classical
logic, but they are not in other logics: for
example, in non-monotonic logic $\Pi' \models \Pi$
may hold, but still $\Pi' \not\equiv \Pi$ even
if $\Pi' \subset \Pi$. In the same way, it can
be that no part of the knowledge base is implied
by the other ones, but still there exists a
proper equivalent subset of it
\cite{libe-redu-3}.

A related definition is that of irredundant equivalent
subset. Such sets result from removing some redundant
clauses while preserving equivalence.

\begin{definition}

A set of clauses $\Pi'$ is said to be an
{\em irredundant equivalent subset} (\ies)
of another set of clauses $\Pi$ if and only
if:

\begin{enumerate}

\item $\Pi' \subseteq \Pi$

\item $\Pi' \equiv \Pi$

\item $\Pi'$ is irredundant

\end{enumerate}

\end{definition}

The second point can be replaced by $\Pi' \models \Pi$
for all monotonic logics.
An alternative definition is that an \ies\  is
an equivalent subset of the original set such
that none of its subsets has the same properties.
Any set of clauses has at least one \ies, but it
may also have more than one of them, as shown by
the following example.

\begin{example}

Let 
$
\Pi = \{ a \vee \neg b, \neg a \vee b ,
	 a \vee c, b \vee c \}
$. This set has two \ies's:\eatpar

\begin{eqnarray*}
\Pi_1 &=& \Pi \backslash \{ a \vee c \} \\
\Pi_2 &=& \Pi \backslash \{ b \vee c \}
\end{eqnarray*}

It is indeed easy to see that the first two clauses
of $\Pi$ are equivalent to $a \equiv b$, which
implies that $a \vee c$ and $b \vee c$ are equivalent.
It is also easy to see that neither $a \vee \neg b$
nor $\neg a \vee b$ can be removed from $\Pi$ while
preserving equivalence with it.

\end{example}

The set of clauses of this example can be used
to show that a set of clauses may have exponentially
many \ies's. Consider the set:

\[
\Pi_n = \bigcup_{i=1,\ldots,n} \Pi[\{a/a_i,b/b_i,c/c_i\}]
\]

In words, $\Pi_n$ is made of $n$ copies of $\Pi$,
each built on its own set of three variables.
While removing clauses from $\Pi_n$, we have $n$
independent choices, one for each copy:
for each $i$ we can remove either $a_i \vee c_i$ or
$b_i \vee c_i$. This proves that $2^n$ outcomes are
possible, each leading to a different \ies

Since a formula may have more than one \ies, its
clauses can be partitioned into three sets: the ones
that are in all \ies's, the ones that are in some
\ies's, and the ones that are in no \ies\  The
idea is that the first clauses are necessary
(they cannot be removed from the set without
changing its semantics), the last ones are
useless (their removal is harmless), while the
other ones are ``useful but not necessary''. We
therefore give the following definitions.

\begin{definition}

A clause $\gamma$ in $\Pi$ is:

\begin{description}

\item[necessary:] it is in all \ies's;

\item[useful:] it is in some \ies's;

\item[useless:] it is not in any \ies.

\end{description}

\end{definition}

Note that useful clauses include all necessary ones,
and that useless and useful are opposite concepts.
In terms of knowledge, necessary clauses express knowledge
in a succinct form, as they are not redundant at all.
Useless clauses can instead be
considered ``strongly redundant'': not only they
can be removed; they can always be removed. In a sense,
they are not saying anything useful from the point of
view of the knowledge they express. On the other hand,
their presence may be important at a meta-level.
For example, the strong redundancy of a clause $\gamma$
may indicate that the information it carries is
very important. It may also indicate
that the piece of knowledge it represents has been
outdated by successive addition, but
further additions to the knowledge base may
require backing up to the part of knowledge we currently
regard as useless. Either way, useless parts may in
some cases be useful at a meta-level. Finally, useful
but not necessary clauses express knowledge that
the knowledge base contains in some other form,
that is, these clauses represent ``one possible way''
of telling this information. As for all redundant
clauses, they may tell that the knowledge they carry
is regarded as important, but they may even indicate
that mistakes has been made in the construction of
the knowledge base, so that two clauses that are believed
to say something different in fact do not.

Technically, checking whether a clause is necessary
is easy, as it does not require cycling over all
possible \ies

\begin{lemma}
\label{necessary-lemma}

A clause $\gamma$ is necessary in $\Pi$ if and only
if $\Pi \backslash \{\gamma\} \not\models \gamma$.

\end{lemma}

\proof If $\Pi \backslash \{\gamma\} \not\models \gamma$,
then $\gamma$ belongs to all \ies's: this is an easy
consequence of the fact that no subset of
$\Pi \backslash \{\gamma\}$ can imply $\gamma$. What
remains to prove is that
$\Pi \backslash \{\gamma\} \models \gamma$ implies that
there is an \ies\ of $\Pi$ that does not contain $\gamma$.
We can build this \ies\   as follows: we start from
$\Pi \backslash \{\gamma\}$ and iteratively remove
clauses that can be derived from it, until we obtain
a set from which no clause can be removed. This is clearly
an \ies, and it does not contain $\gamma$.~\qed

Checking inutility of a clause cannot be expressed
with a simple condition like this one. This is shown
by Theorem~\ref{complexity-useful}, which proves that
the opposite problem of telling whether a clause is
useful is \P{2}-complete. As a result, the definition
of uselessness cannot be expressed as a single entailment
check (like the one for necessity) unless exponentially
large formulae are used.

Any set of clauses has at least one \ies. Checking the
existence of an \ies\  is thus trivial. On the other hand,
a set may have more than one \ies. Deciding uniqueness of
\ies's for a specific set of clauses is important, as it
tells whether there is a choice among the possible minimal
representations of the same piece of information. For
example, a trivial algorithm for producing an \ies\  is
that of iteratively removing the first clause
that is implied by the other ones. This algorithm clearly
outputs an \ies. However, other ones may exist, and
be better either because are shorter (have
less clauses), or because their structure make them more
effective to use (for example, they are Horn or in a similar
special form that makes reasoning with them easier.)

This problem is also of interest because uniqueness
implies that all clauses are either necessary or useless.
As a result, checking usefulness and inutility becomes
the same and opposite problem of necessity, respectively.
Therefore, they become much simpler than in the general
case.

Clearly, if a set is irredundant, it has a single \ies\
On the other hand, some sets may be redundant but have
a single \ies\  anyway. The following example shows such
a set.

\[
\Pi = \{a \vee b, a \vee \neg b, a \vee c\}
\]

The first two clauses are in fact equivalent to $a$,
which makes $a \vee c$ redundant. On the other hand,
$a \vee c$ cannot be used to infer $a$. As a result,
the only \ies\  of this set is composed of its first
two clauses.

The condition of uniqueness is formally defined as:
there exists exactly one $\Pi'$ that is a subset of
$\Pi$ and is irredundant. However, the following lemma
shows an easier why to determine whether a set of
clauses has a single \ies

\begin{lemma}
\label{unique-lemma}

A set of clauses $\Pi$ has a unique \ies\  if
and only if $\Pi_N \equiv \Pi$, where $\Pi_N$ is
the set of necessary clauses:

\[
\Pi_N = \{ \gamma \in \Pi ~|~
\Pi \backslash \{\gamma\} \not\models \gamma \}
\]

\end{lemma}

\proof If $\Pi$ has a unique \ies, then
its clauses are exactly the clauses that
are in all \ies's of $\Pi$. Lemma~\ref{necessary-lemma}
tells that the clauses that are contained in all
\ies's can be expressed as $\Pi_N$.

Let us now assume that $\Pi_N \models \Pi$,
and prove that $\Pi_N$ is the unique \ies\  of $\Pi$.
Since no clause of $\Pi_N$ is implied by the rest
of $\Pi$, it is not implied by the rest of $\Pi_N$
either, which proves that $\Pi_N$ is an \ies\  We
only have to prove that $\Pi$ does not have any other
\ies. Assume, by contradiction, that $\Pi' \not= \Pi_N$
is an \ies: if $\Pi_N \subset \Pi'$, then $\Pi'$
is not irredundant; otherwise, there exists
$\gamma \in \Pi_N \backslash \Pi'$. This condition
can be decomposed into $\gamma \in \Pi_N$ and
$\gamma \not\in \Pi'$. The first formula implies
$\Pi \backslash \{\gamma\} \not\models \gamma$ since
$\Pi_N$ is the set of necessary clauses. The second
formula, together with $\Pi' \models \Pi$, implies
that $\Pi' \backslash \{\gamma\} \models \gamma$.
This is a contradiction, as $\Pi' \subseteq \Pi$.~\qed

The following condition is sufficient
for proving that a set of clauses has more than one
\ies\  Intuitively, while removing redundant clauses
from a set, we may arrive to a point in which we have
a choice to make between removing one clause in a
pair. If this is the case, this choice produces two
different \ies's.

\begin{lemma}
\label{more-lemma}

If $\Pi \backslash \{\gamma_1\} \equiv \Pi$
and $\Pi \backslash \{\gamma_2\} \equiv \Pi$
but $\Pi \backslash \{\gamma_1, \gamma_2\} \not\equiv \Pi$,
then $\Pi$ has at least two \ies's

\end{lemma}

\proof Since any set of clauses has at least
an \ies, the same happens for $\Pi \backslash \{\gamma_1\}$.
Let therefore $\Pi'$ be an \ies\  of
$\Pi \backslash \{\gamma_1\}$. Since
$\Pi \backslash \{\gamma_1\}$ is equivalent
to $\Pi$, this is also an \ies\  of $\Pi$.
It does not contain $\gamma_1$ because it
is a subset of $\Pi \backslash \{\gamma_1\}$.
We show that it necessarily contains $\gamma_2$.
Suppose it does not: then
$\Pi' \subset \Pi \backslash \{\gamma_1, \gamma_2\}$
which, by assumption, is not equivalent to $\Pi$,
contrarily to the claim that it is.

We have therefore proved that any \ies\
of $\Pi \backslash \{\gamma_1\}$ is an \ies\
of $\Pi$ that contains $\gamma_2$ but not
$\gamma_1$. For the same reasons, any
\ies\  of $\Pi \backslash \{\gamma_2\}$
is an \ies of $\Pi$ that contains $\gamma_1$
but not $\gamma_2$. As a result, the \ies's
of $\Pi \backslash \{\gamma_1\}$ and
$\Pi \backslash \{\gamma_2\}$ are all different.
Since each set has at least an \ies, we have
proved that $\Pi$ has at least two \ies's.~\qed

This condition is however not necessary.
Indeed, the choice between {\em two} clauses
may show up only when some redundant clauses
have already been removed. This happens for
example when two clauses out of three have
to be removed, like in the following set:

\[
\Pi = \{ a \equiv b ,~ a \equiv c ,~
a \vee d ,~ b \vee d, ~ c \vee d \}
\]

The clauses composing the first two formulae
are necessary. Since $a$, $b$, and $c$
are equivalent, and the last three
clauses are equivalent as well. As a result,
we can always remove two of them. This proves
that the condition of the theorem (which
requires two non-necessary clauses not to
be removable at the same time) is false,
while the set has more than one \ies.

\

Let us now show some properties that will be useful for
the complexity analysis of problems related to redundancy
and \ies's. Checking whether a specific clause is redundant
is easy to characterize from a computational point of view,
as it amounts to exactly one entailment test:
$\Pi \backslash \{\gamma\} \models \gamma$. On the
other hand, results about the redundancy of a whole
set are harder to obtain, as we have to make sure
that the clauses we define do not interact to form
redundancy when they should not. In other words,
we can still use the fact that proving that
$\Pi \backslash \{\gamma\} \models \gamma$ is
\conp-complete, but this is a real reduction only
if $\Pi$ does not contain any other redundant clauses.
The hardness proofs in this paper are indeed based
on the following method:

\begin{quote}

the formula resulting from a reduction contains parts
that are known to be irredundant.

\end{quote}

These parts will be then useful, because they express
some constraints on the model, while they do not
affect redundancy. Since the only parts that are
``known to be irredundant'' are the necessary clauses,
this method can be used for problems about \ies's
as well. The following definition shows how
clauses can be made irredundant.

\begin{definition}

The irredundant version of a set of clauses
$\Gamma = \{ \gamma_1, \ldots, \gamma_m \}$
is defined as:

\[
\Gamma[C] = \{ c_i \rightarrow \gamma_i ~|~ \gamma_i \in \Gamma \}
\]

\noindent where $C=\{c_1,\ldots,c_m\}$ are variables
of the same number of the clauses of $\Gamma$
($c_i \rightarrow \gamma_i$ denotes the clause
$\neg c_i \vee \gamma_i$.)

\end{definition}

The point of this definition is that $\Gamma[C]$ is
composed of necessary clauses only. The following
lemma shows exactly how this can be proved.

\begin{lemma}
\label{irredundant-lemma}

For any set of clauses $\Gamma$ containing no
tautologies, the model $\omega_i$ below satisfies
all clauses of $\Gamma[C]$ but
$c_i \rightarrow \gamma_i$:

\[
\omega_i(\Gamma, C) = \{ c_i \} \cup
\{ \neg l_j ~|~ l_j \in \gamma_i \}
\]

\end{lemma}

\proof $\omega_i(\Gamma, C)$ is not a model of $c_i \rightarrow \gamma_i$,
as we assumed that no clause is tautological. On the converse,
it is a model of $\Gamma[C] \backslash \{ c_i \rightarrow \gamma_i \}$
simply because it falsifies all $c_j$'s with $j \not=i$.~\qed

This lemma actually proves that all clauses of
$\Gamma[C]$ are irredundant. We do not state the
lemma this way because the reductions use $\Gamma[C]$
in conjuction with other clauses: in order to prove
that the clauses of $\Gamma[C]$ are irredundant, we
extend the models $\omega_i(\Gamma, C)$ in such a way they
satisfy all other clauses.

While it is simple to prove that $\Gamma$ is unsatisfiable
if and only if
$\Gamma[C] \models \neg c_1 \vee \cdots \vee \neg c_m$, we
cannot simply add this clause to $\Gamma[C]$ to show
the hardness of the irredundancy problem, as this
clause may make some clauses of $\Gamma[C]$ redundant. The
complete proof requires adding a new variable to that
clause to avoid this problem.

\begin{lemma}
\label{sat-lemma}

For any set of clauses $\Gamma$, none of the clauses
of $\Gamma[C]$ is redundant in $\Gamma[C,a]$ below:

\[
\Gamma[C,a] = \Gamma[C] \cup
\{\neg c_1 \vee \cdots \vee \neg c_m \vee \neg a\}
\]

\noindent where $a$ is a new variable, while the clause
$\neg c_1 \vee \cdots \vee \neg c_m \vee \neg a$ is
redundant (\ie,
$\Gamma[C] \models \neg c_1 \vee \cdots \vee \neg c_m \vee \neg a$)
if and only if $\Gamma$ is unsatisfiable.

\end{lemma}

\proof Lemma~\ref{irredundant-lemma} proves that
$\omega_i[C]$ is a model of all clauses of $\Gamma[C]$ but
$c_i \rightarrow \gamma_i$. Since it is also a model
of the last clause ($a$ is implicitly assumed to be
false in $\omega_i[C]$), no clause of $\Gamma[C]$ is implied
by the other ones. Let us now prove that the redundancy of
the last clause is related to the satisfiability
of $\Gamma$.

\begin{description}

\item[$\Gamma$ is unsatisfiable.] Since $\Gamma$ has
no models, no model of $\Gamma[C]$ contains all $c_i$'s.
As a result, $\Gamma[C] \models \neg c_1 \vee \cdots \vee \neg c_m$,
which implies that
$\Gamma[C] \models \neg c_1 \vee \cdots \vee \neg c_m \vee \neg a$,
which in turns implies that $\Gamma[C,a]$ is redundant.

\item[$\Gamma$ is satisfiable.] We prove that the
last clause of $\Gamma[C,a]$ is irredundant (the other
ones have already proved to be so.) Since $\Gamma$ is
satisfiable, it has a model $\omega$. By setting all $c_i$'s
and $a$ to be true, we obtain the model
$\omega \cup \{c_1,\ldots,c_m,a\}$, which satisfy $\Gamma[C]$.
This is not a model of $\neg c_1 \vee \cdots \vee \neg
c_m \vee \neg a$: as a result, the last clause is
irredundant.

\end{description}~\qed

This lemma is used not only in the proof of hardness
of the problem of checking redundancy, but also for
the proof of hardness of other problems (such as
checking necessity of a clause.)

\section{Complexity Results}

In this section, we show the complexity results
that are summarized in Table~\ref{complexity-table}.
The first result is about the complexity of checking
whether a set of clauses is redundant.

\begin{theorem}
\label{th-irre}

Checking irredundancy of a set of clauses is \np-complete.

\end{theorem}

\proof Membership: we have to check whether,
for any $\gamma \in \Pi$, it holds
$\Pi \backslash \{ \gamma \} \not\models \gamma$.
This can be done by guessing a model for each set
$\Pi \backslash \{ \gamma \} \cup \{\neg \gamma\}$,
which shows the problem to be in \np.

Hardness is an easy consequence of
Lemma~\ref{sat-lemma}: a non-tautological
set of clauses $\Gamma$ is satisfiable if and only
if $\Gamma[C,a]$ is irredundant.~\qed

What this theorem proves is that checking irredundancy
of a set of clauses is not harder, theoretically, than
checking whether a single clause is irredundant. 
Although the problem looks harder than entailment,
it is indeed the hardness
proof the more complex part of the completeness proof
(it requires using Lemma~\ref{sat-lemma}, which in
turns requires Lemma~\ref{irredundant-lemma}.)
This is because redundancy does not immediately allow
expressing entailment (the irredundant version of a set
of clauses has been introduced exactly for solving this
problem.)

Let us now turn to the problems related to \ies's.
The first problem is that of checking whether a set
of clauses is an \ies\  of another one. This problem
clearly requires checking equivalence and irredundancy.
The following theorem actually proves that the
problem is hard for the class \Dp, which contains
all problems that can be decomposed into a problem
in \np\  and a problem in \conp.

\begin{theorem}

Given two sets of clauses $\Pi$ and $\Pi'$, checking
whether $\Pi'$ is an \ies\  of $\Pi$ is \Dp-complete.

\end{theorem}

\proof Membership amounts to showing that $\Pi' \subseteq \Pi$
(a polynomial task), that $\Pi' \models \Pi$ (which is
in \conp) and that $\Pi'$ is irredundant (which we proved
to be in \np). Therefore, the problem is in \Dp.

Hardness is proved by reduction from the \sat-\unsat\
problem: given a pair of sets of clauses
$\l \Gamma, \Sigma \r$, check whether the first one is
satisfiable while the second one is not. This problem is
\Dp-complete even if $\Gamma$ and $\Sigma$ do not share
variables \cite{blas-gure-82}, which we assume. Let $C$
and $D$ be new sets of variables in one-to-one correspondence
with the clauses of $\Gamma$ and $\Sigma$, respectively.
Let $a$ and $e$ be two other new variables.
Reduction is as follows:\eatpar

\begin{eqnarray*}
\Pi &=& \Gamma[C,a] \cup \Sigma[D,e] \\
\Pi' &=& \Gamma[C,a] \cup \Sigma[D]
\end{eqnarray*}

First, we show that $\Pi'$ is irredundant if and only
if $\Gamma$ is satisfiable. By Lemma~\ref{irredundant-lemma},
$\Sigma[D]$ is irredundant. Lemma~\ref{sat-lemma} proves
that $\Gamma[C,a]$ 
is irredundant if and only if $\Gamma$ is satisfiable.
Since these two subsets of $\Pi'$ do not share variables,
$\Pi'$ is irredundant if and only if both parts are,
that is, $\Pi'$ is irredundant if and only if $\Gamma$
is satisfiable.

What remains to prove is only that $\Pi' \models \Pi$ if and
only if $\Sigma$ is unsatisfiable. By Lemma~\ref{sat-lemma},
$\Sigma[D] \models \neg d_1 \vee \cdots \vee \neg d_r \vee \neg e$
holds if and only if $\Sigma$ is unsatisfiable.~\qed

Given that a set of clauses can have more than
one \ies, it is of interest to check the size
of minimal \ies's, as it tells the amount of
redundant information the theory contains, and
also how much the size of the knowledge base
can be reduced by deleting redundant clauses.
The decision problem we consider is that of
checking the existence of \ies's of size bounded
by a constant integer. This problem can be solved
by iterating over all possible \ies's. Such
a procedure amounts to checking whether there
exists a subset that is a \ies\  and has the
given size. The following theorem tells that this
iteration cannot be avoided in general.

\begin{theorem}
\label{exists-size}

Given a set of clauses $\Pi$ and an integer
$k$, deciding whether $\Pi$ has an \ies\
of size at most $k$ is \S{2}-complete.

\end{theorem}

\proof Membership: the problem amounts to deciding
whether there exists a subset of $\Pi$ that is
equivalent to it and of size at most $k$. Since
the problem can be expressed as a $\exists\forall$QBF,
it is in \S{2}.

Hardness is proved via a quite complicated reduction
from $\exists\forall$QBF. Let $\exists X \forall Y. \neg \Gamma$
be a formula, where
$\Gamma = \{ \gamma_1, \ldots, \gamma_m \}$ is a set of
clauses. This problem is \S{2}-hard, as it
is the complement of the problem of deciding whether
a $\forall\exists$QBF, in which the matrix is a CNF
formula, is valid \cite{stoc-meye-73}.
We build a set $\Pi$ as the union of the following
sets of clauses:

\begin{eqnarray*}
\Pi_1 &=& \bigcup^{j=1,\ldots,r}_{i=1,\ldots,n}
          \{ x^j_i, z^j_i \}
\\
\Pi_2 &=& \bigcup_{i=1,\ldots,n}
          \{ x^1_i \wedge \cdots \wedge x^r_i \rightarrow x_i,
             z^1_i \wedge \cdots \wedge z^r_i \rightarrow z_i \}
\\
\Pi_3 &=& \bigcup_{i=1,\ldots,n}
          \{ x_i \rightarrow w_i,
             z_i \rightarrow w_i \}
\\
\Pi_4 &=& \bigcup_{j=1,\ldots,m}
          \{ w_1 \wedge \cdots \wedge w_n \rightarrow \gamma_j^N \}
\\
\Pi_5 &=& \{ v_1, \ldots, v_t, \neg v_1 \vee \cdots \vee \neg v_t \}
\end{eqnarray*}

Here, $\gamma^N_j$ is obtained from $\gamma_j$ by replacing
every positive occurrence of $x_i$ with $\neg z_i$. The values
of the constant $k$, $r$, and $t$ are chosen as follows:
if $n$ is the number of variables and $m$ the number of
clauses of $\Gamma$, we set $r=m+1$, $k=(r+2) \cdot n +m$,
and $t=k+1$.

We prove that $\exists X \forall Y. \neg \Gamma$ is valid
if and only if 
$\Pi = \Pi_1 \cup \Pi_2 \cup \Pi_3 \cup \Pi_4 \cup \Pi_5$
has an equivalent subset of size at most $k$.

The set $\Pi$ is unsatisfiable because $\Pi_5$
is unsatisfiable. Therefore, we are looking for a
subset of $\Pi$ of size at most $k$ that is unsatisfiable.
Note that, removing even a single clause from $\Pi_5$, it
becomes satisfiable. Since $\Pi_5$ does not share any variable
with the other subsets, it follows that no proper subset
of $\Pi_5$ can contribute to the generation of unsatisfiability.
Since $t>k$, if an unsatisfiable subset of size less than
$k$ contains clauses from $\Pi_5$, they can be removed
while maintaining unsatisfiability. As a result,
while looking for an unsatisfiable subset of $\Pi$,
clauses of $\Pi_5$ can be disregarded: these clauses
are only used to guarantee that $\Pi$ is unsatisfiable.

We have therefore proved that $\Pi$ has an \ies\  of size
bounded by $k$ if and only if 
$\Pi_1 \cup \Pi_2 \cup \Pi_3 \cup \Pi_4$ has an inconsistent
subset of size bounded by $k$. Let us therefore consider
$\Pi' \subseteq \Pi_1 \cup \Pi_2 \cup \Pi_3$ and
$\Pi'' \subseteq \Pi_4$, and see what happens when
$\Pi' \cup \Pi''$ is an unsatisfiable set of at most
$k$ clauses.

First, neither $\Pi'$ nor $\Pi''$ is unsatisfiable alone,
as both $\Pi_1 \cup \Pi_2 \cup \Pi_3$ and $\Pi_4$ are
satisfiable (the first is satisfied by the model that
evaluates to true all variables, the second by the model
that evaluates to false all variables.)
Second, if $\Pi'$ does not imply all $w_i$'s, then
$\Pi' \cup \Pi_4$ is satisfiable, and therefore
$\Pi' \cup \Pi''$ is satisfiable as well. There exists
exactly two minimal subsets of $\Pi_1 \cup \Pi_2 \cup \Pi_3$
that imply $w_i$:

\begin{eqnarray*}
\Sigma_i
&=&
\bigcup_{j=1,\ldots,r} \{ x^j_i \} \cup
\{ x^1_1 \wedge \cdots \wedge x^r_i \rightarrow x_i, x_i \rightarrow w_i \}
\\
\Sigma'_i
&=&
\bigcup_{j=1,\ldots,r} \{ z^j_i \} \cup
\{ z^1_1 \wedge \cdots \wedge z^r_i \rightarrow z_i, z_i \rightarrow w_i \}
\end{eqnarray*}

These two sets have the same size. The number $k$
has been chosen so that $k=n \cdot (r+2) +m = n \cdot |\Sigma_i| + m$.
Since all $w_i$'s have to be implied, $\Sigma_i \subset \Pi'$
or $\Sigma_i' \subset \Pi'$ for each $i$. Since $m<|\Sigma_i|$, we have
that $k< n \cdot (|\Sigma_i|+1)$, that is, $\Pi'$ cannot contain
more than $n$ sets $\Sigma_i$ or $\Sigma_i'$. More precisely,
$r+1=m+2$ other clauses are necessary to imply another $x_i$
or $z_i$, which are shared with $\Pi''$. Therefore, $\Pi'$
must contain exactly one group among $\Sigma_i$ and $\Sigma'_i$
for any $i$, which amounts to $n \cdot (r+2)$ clauses. The
remaining $m$ clauses can be taken from $\Pi''$. Since
$\Pi_4$ has size $m$, we can simply take $\Pi''=\Pi_4$.

We have proved that $\Pi'$ implies either
$x_i$ or $z_i$, for any $i$, but not both. Candidate
unsatisfiable subsets are therefore in correspondence with
truth assignments on the variables $x_i$. Moreover, all
variables $w_i$ are true, which makes $\Pi_4$ equivalent
to $\bigcup_{j=1,\ldots,m} \{ \gamma^N_j \}$. If $\Pi'$
contains $x_i$, then $\neg x_i$ can be removed from any
clause $\gamma^N_j$ containing it, while $\neg z_i$ remains.
The opposite happens if $z_i$ is in $\Pi'$.

Either way, if a variable of $\{x_i,z_i\}$ is in $\Pi'$,
the other one is not mentioned in $\Pi'$, so we can assign
it to false in order to satisfy as many clauses as possible
(we are trying to prove unsatisfiability, so we have to test
the most unfavorable possibility). What remains of $\Pi_4$
is the set $\Gamma$ in which all variables $x_i$ has been
removed, by assigning them either to true (if
$\Sigma_i \subset \Pi'$) or to false
(if $\Sigma'_i \subset \Pi'$). Therefore, the choice of
including $\Sigma_i$ or $\Sigma_i'$ makes $\Pi_4$ equivalent
to $\Gamma$ after setting $x_i$ to some truth value.
Therefore, $\Pi$ has an unsatisfiable subset of size $k$
if and only if $\exists X \forall Y . \neg \Gamma$ is true.

Note that the choice of
an unsatisfiable set $\Pi$ is not necessary. Indeed,
by adding a new variable $u$ to all clauses, $\Pi$ and
all its subsets are made satisfiable. Since $\Pi$ is
now equivalent to $u$, one of its subsets can be
equivalent to it only if, assigning false to $u$, leads
to unsatisfiability, which has been proved to be
equivalent to the QBF problem.~\qed

This theorem implies that, unless the polynomial
hierarchy collapses, the problem of checking
the existence of \ies's of size bounded by $k$
is not in any class below \S{2}. As a result,
the definition of the problem is not equivalent
to a condition that contains ``less quantifiers'',
unless an exponential blow-up is introduced.
In other words, any condition that do not require
checking exponentially sized formulae will contain
an initial part ``there exists something...''
similar to the part ``there exists $\Pi'$...'' of
the original definition. Such a simpler equivalent
condition would indeed imply that the problem is
in \np\  or \conp.

This is not the case for the problem of checking
the membership of a clause to all \ies's, on the
other hand: while the initial definition is
``for all $\Pi' \subseteq \Pi$,...'', we proved
it equivalent to
$\Pi \backslash \{\gamma\} \not\models \gamma$.
This simplification is however only possible because
the problem is easier than what appears from the definition:
while the definition of the problem can be expressed
as a $\forall\exists$QBF (implying that the problem
is in \P{2},) Lemma~\ref{necessary-lemma} proved that
the problem is actually in \conp. The next theorem
also shows that the problem is hard for that class
(and cannot therefore be simplified to a condition
that do not require a satisfiability/entailment
test at all.)

\begin{theorem}

Deciding whether a clause is necessary in a set
(it is contained in all its \ies's) is \np-complete.

\end{theorem}

\proof By Lemma~\ref{necessary-lemma}, a clause
is necessary if and only if
$\Pi \backslash \{\gamma\} \not\models \gamma$,
and this problem is in \np. Hardness easily
follows from Lemma~\ref{sat-lemma}: since all
clauses of $\Gamma[C,a]$ are irredundant but
(possibly) the last one, $\Gamma[C,a]$ has exactly
one \ies, which is either $\Gamma[C]$ or $\Gamma[C,a]$,
depending on the satisfiability of $\Gamma$.
As a result, the only clause of
$\Gamma[C,a] \backslash \Gamma[C]$ is in all
\ies's if and only if $\Gamma$ is satisfiable.~\qed

While deciding whether a clause is in all \ies's is
in \np, the similar problem of deciding whether a
clause is in {\em at least one} \ies\  is complete
for the class \S{2}, and is therefore harder. This result
is somehow surprising, as these two problems have
very similar definitions, and checking the existence
of an \ies\  containing a clause may look even
simpler than checking all of them.

\begin{theorem}
\label{complexity-useful}

Deciding whether a clause $\gamma$ is in at least one \ies\
of a set of clauses $\Pi$ is \S{2}-complete.

\end{theorem}

\proof Membership is trivial: the problem can be expressed
as the existence of a set $\Pi' \subseteq \Pi$ containing
$\gamma$ that is equivalent to $\Pi$ and irredundant.

Hardness is proved by reduction from $\exists\forall$QBF.
We assume that the matrix of the QBF formula is the negation
of a CNF: this problem is \S{2}-hard, as it is the complement
of deciding whether a $\forall\exists$QBF formula, in which
the matrix is in CNF, is valid \cite{stoc-meye-73}.
We prove that $\exists X \forall Y . \neg \Gamma$ is valid
(where $\Gamma=\{\gamma_1,\ldots,\gamma_m\}$)
if and only if $w$ is in at least one \ies\  of the
following set $\Pi$:

\[
\Pi =
\bigcup_{i=1,\ldots,n} \{ x_i, \neg x_i \} \cup
\{ w \} \cup
\bigcup_{i=1,\ldots,m} \{ w \rightarrow \gamma_i \}
\]

This set is clearly unsatisfiable. Its \ies's are its
unsatisfiable minimal subsets. Let us now show how a
subset $\Pi'$ of this kind is composed. If both $x_i$ and
$\neg x_i$ are in $\Pi'$, they are enough to generate
contradiction, so no other clause can be in $\Pi'$,
otherwise the other clauses would be redundant. We have
therefore found a first group of minimal unsatisfiable
subsets of $\Pi$: those composed exactly of a pair
$\{x_i, \neg x_i\}$.

Let us now try to build an unsatisfiable $\Pi' \subseteq \Pi$
that contains $w$. Besides $w$, such set $\Pi'$ can include
$\bigcup_{i=1,\ldots,m} \{ w \rightarrow \gamma_i \}$,
as well as a literal between $x_i$ and $\neg x_i$ for
any $i$ (but not both, otherwise the other clauses
would be redundant). It is now evident that such set can
be unsatisfiable only if, for the given choice of the
$x_i$'s, the set $\Gamma$ is unsatisfiable. Thus,
there exists an unsatisfiable subset of $\Pi$ containing
$w$ if and only if $\Gamma$ is unsatisfiable. What remains
to prove is that any \ies\  obtained by removing redundant
clauses from $\Pi'$ contains $w$, but this is an easy
consequence of the fact that $\Pi' \backslash \{w\}$ is
satisfiable.~\qed

The hardness result proves that, unlike the necessary
condition, the definition of usefulness cannot be reduced
to a simple entailment/satisfiability check, unless the
polynomial hierarchy collapses or some exponentially
large formulae are used.

The problem of uniqueness amounts to checking whether a
set of clauses has a single \ies\  This problem can be
solved without cycling over all possible subsets of
clauses, as Lemma~\ref{unique-lemma} proves that finding
the set of necessary clauses suffices.

\begin{theorem}

Deciding whether a set of clauses $\Pi$ has a single
\ies\  is \Dlog{2}\  complete.

\end{theorem}

\proof By Lemma~\ref{unique-lemma}, all we have to
do is to check whether the set of necessary clauses
$\Pi_N$ is equivalent to $\Pi$. In turns, the set
of necessary clauses can be found by checking
$\Pi \backslash \{\gamma\} \not\models \gamma$ for
each clause $\gamma \in \Pi$. As a result, we perform
a polynomial number of parallel calls to an oracle in
\np\  (each one to check whether a clause is necessary)
followed by a single other call (to check equivalence
between $\Pi_N$ and $\Pi$.) By a well-known result by
Gottlob \cite{gott-95-jacm}, the problem is in \Dlog{2}.

We prove that the problem of uniqueness is
\Dlog{2}-hard by reduction from the problem
of odd satisfiability: given a sequence of
sets of clauses $(\Pi^1, \ldots, \Pi^r)$,
each built on its own alphabet,
such that the unsatisfiability of $\Pi^j$
implies that of $\Pi^{j+1}$, decide
whether the first $\Pi^k$ that is unsatisfiable
is of odd index, that is, $k$ is odd.

For each set of clauses $\Pi^j$, we need an additional
set of variables $C^j=\{c^j_1,\ldots,c^j_m\}$ and three
other variables $a^j$, $b^j$, and $c^j$. We define
$\gamma^j_g =\neg c^j_1 \vee \cdots \vee \neg c^j_m$.
As proved by Lemma~\ref{sat-lemma}, $\Pi^j[C^j]$ implies
$\gamma^j_g \vee d$, where $d$ is a variable not occurring
in $\Pi^j[C^j]$ (\eg, $a^j$), if and only if $\Pi^j$ is
unsatisfiable.

Let $j$ be an odd index between $1$ and $r$. Define:\eatpar

\begin{eqnarray*}
\Pi^j_D
&=&
\Pi^j[C^j] \cup
\{ \gamma^j_g \vee a^j \vee c^j,~
   \gamma^j_g \vee b^j \vee c^j \}
\\
\Pi^{j+1}_D
&=&
\Pi^{j+1}[C^{j+1}] \cup
\{ \gamma^j_g \vee c^{j+1}_i \vee a^j \vee \neg b^j ~|~
\gamma^{j+1}_i \in \Pi^{j+1} \} \cup
\\
&&
\{ \gamma^j_g \vee c^{j+1}_i \vee \neg a^j \vee b^j ~|~
\gamma^{j+1}_i \in \Pi^{j+1} \}
\end{eqnarray*}

Variables are only shared between $\Pi^j_D$
and $\Pi^{j+1}_D$, and only if $j$ is odd.
We therefore only have to check whether
$\Pi^j_D \cup \Pi^{j+1}_D$ has a unique \ies,
where $j$ is odd.

Let us consider the easiest cases first.
By Lemma~\ref{sat-lemma}, if $\Pi^j$ is unsatisfiable,
then the clauses $\gamma^j_g \vee c^j$ and
$\gamma^j_g \vee c^{j+1}$ are entailed by
$\Pi^j[C^j]$. As a result, all clauses but
those in $\Pi^j_D \cup \Pi^{j+1}_D$ are redundant.
Since the clauses in $\Pi^j_D \cup \Pi^{j+1}_D$ are
irredundant, we have a single \ies\  $\Pi^j_D \cup \Pi^{j+1}_D$.

The second easy case is when $\Pi^{j+1}_D$ unsatisfiable.
By Lemma~\ref{sat-lemma}, $\Pi^{j+1}[C^{j+1}]$ implies
$\neg c^{j+1}_1 \vee \cdots \neg c^{j+1}_m$, that is,
at least a variable $c^{j+1}_i$ is false. As a result,
$(a^j \equiv b^j) \vee \gamma^j_g$ is entailed.
Therefore, the two last clauses of $\Pi^j_D$
are made equivalent; therefore, one of them
can be removed, but not both. By Lemma~\ref{more-lemma},
$\Pi$ has have more than one \ies\

The longest part of the proof is to prove that,
if both $\Pi^j$ and $\Pi^{j+1}$ are satisfiable,
then all clauses are irredundant. This is proved
by showing, for each clause, a model of the other
clauses that is not a model of it. For the clauses
in $\Pi^j[C^j]$ this is the model
$\omega_i(C^j)$ of Lemma~\ref{irredundant-lemma},
extended by setting all $C^j$ to false, and $c^j$,
$a^j$, and $b^j$ to true.

For the clause $a^j \vee c^j \vee \gamma^j_g$, we
choose the model evaluating all $c^j_i$ and $c^{j+1}_i$
to true, all variables of $\Pi^j$ and $\Pi^{j+1}$
according to their respective models, both $a^j$ and
$c^j$ to false, and $b^j$ to true. This model does
not satisfy $a^j \vee c^j \vee \gamma^j_g$ by
construction, but satisfies all other clauses. Indeed,
all clauses of $\Pi^j[C^j] \cup \Pi^{j+1}[C^{j+1}]$ are satisfied
because we have chosen the models of $\Pi^j$ and $\Pi^{j+1}$,
the clause $b^j \vee c^j \vee \gamma^j_g$ is satisfied
because of $b^j$, and the clauses
$c^{j+1}_i \vee [\neg] a^j \vee [\neg] b^j \vee \gamma^j_g$
are satisfied because of $c^{j+1}_i$. For the clause
$b^j \vee c^j \vee \gamma^j_g$, the model with the
values of $a^j$ and $b^j$ swapped works in the same way.

The clause $c^{j+1}_i \vee a^j \vee \neg b^j \vee \gamma^j_g$
is falsified by the model that evaluates $c^{j+1}_i$
to false, $a^j$ to false, $b^j$ to true, $c^j$ to
true, all $c^j_i$ to true, all $c^{j+1}_z$ with
$z \not= i$ to true, and the variables of
$\Pi^j$ and $\Pi^{j+1}$ according to their respective
models. This model satisfies all other clauses:
indeed, the clauses of $\Pi^j[C^j] \cup \Pi^{j+1}[C^{j+1}]$ 
are satisfied by the choice of the variables of
$\Pi^j$ and $\Pi^{j+1}$; all clauses with $b^j$ or
$c^j$ are satisfied as well; the only remaining
clauses are those of the form
$c^{j+1}_z \vee a^j \vee \neg b^j \vee \gamma^j_g$,
with $z \not= i$, which are however satisfied by
the truth value of $c^{j+1}_z$.~\qed

\section{Query Equivalence}

The definition of equivalence that is most commonly
used is that of logical equivalence: two formulae
are equivalent if and only if they have the same sets
of models. This definition is the same as the following
one.

\begin{quote}

two formulae $\Pi_1$ and $\Pi_2$ are logically
equivalent if and only if, for any formula $\Gamma$,
it holds $\Pi_1 \models \Gamma$ if and only if
$\Pi_2 \models \Gamma$.

\end{quote}

This definition is formally equivalent to the previous
one, but emphasizes a common use of propositional formulae:
if a formula $\Pi_1$ represents a piece of knowledge,
reasoning is usually (but not always) done in terms of
queries. In turns, querying a knowledge base means
checking whether some facts follow from it or not.
Formally, given a piece of knowledge represented by
a formula $\Pi_1$, querying it means checking whether
a fact represented by another formula $\Gamma$ follows
from it, that is, whether $\Pi_1 \models \Gamma$. If
the above condition on $\Pi_1$ and $\Pi_2$ holds, we
can say that $\Pi_2$ represents the same knowledge as
$\Pi_1$ as these two formulae are
indistinguishable from the point of view of reasoning.

This new definition of equivalence is of interest because
it can be extended in many directions. Namely, if not all
formulae are possible queries, it
does not coincide any more with logical equivalence. Two
cases have been considered in the past:

\begin{enumerate}

\item we are only interested in queries
that are in a particular syntactic form, for example,
the Horn form~\cite{cado-doni-97};

\item we are only interested in formulae about a
subset of variables
\cite{cado-etal-97-c,lang-etal-03}.

\end{enumerate}

On the other hand, we may also interested in a set
of queries that {\em strictly include} the
set of propositional formulae. This is the case, for
example, when queries can be conditional formulae
like $\Gamma > \Sigma$, which means ``if $\Gamma$ were
true, would $\Sigma$ holds?''

\begin{enumerate}

\setcounter{enumi}{2}

\item we are interested into all possible
conditional queries \cite{libe-zhao-03}.

\end{enumerate}

Intuitively, $\Gamma > \Sigma$ is entailed by $\Pi$
if and only if $\Sigma$ follows from the formula
that is obtained by revising $\Pi$ with $\Gamma$.
This motivates this kind of equivalence: two
formulae are equivalent if and only if they are
logically equivalent, and remain so regardless of updates.
This kind of equivalence is related to strong equivalence
in logic programming \cite{lifs-pear-valv-01}, and
has been defined for propositional logic by
Liberatore and Zhao~\cite{libe-zhao-03}.

We call any form of equivalence that is based on a
particular set of consequences {\em query equivalence}
(this name has been used by Cadoli et al.
\cite{cado-etal-97-c} for the definition based on
a subset of variables, but it is somehow inappropriate
as other sets of queries make sense.) The two
forms of equivalence above (based on considering subsets
of propositional formulae) are called Horn equivalence
and var-equivalence, respectively. The form of equivalence
based on conditional statements is instead called
strong equivalence or conditional equivalence.

Since redundancy is defined in terms of equivalence
(a set is redundant if and only if it is equivalent
to a proper subset of its,) using a definition of
equivalence that is different from the logical one
leads to different properties and results.
Using query equivalence, redundancy
tells which clauses are really necessary \wrt\  a given
set of queries. We only consider two kinds of equivalence:
var-equivalence and conditional equivalence. The two
corresponding forms of redundancy are called
var-redundancy and conditional redundancy.

\let\subsectionnewpage=\newpage
\subsection{Var-Redundancy}

Var redundancy is defined in the same way as logical
redundancy, but using var-equivalence instead of
logical equivalence. This kind of equivalence
is called {\em query equivalence} by Cadoli et al.
\cite{cado-etal-97-c} and {\em var-equivalence} by
Lang, Liberatore, and Marquis \cite{lang-etal-03}.
We prefer the second name, and reserve the first one
for the more general concept of equivalence based on
an arbitrary set of queries. Formally, var-equivalence
is defined as follows.

\begin{definition}[Var-Equivalence \cite{lang-etal-03}]

Two formulae $\Pi_1$ and $\Pi_2$ are var-equivalent
\wrt\ a set of variables $V$ if and only if, for
each formula $\Gamma$ over variables $V$, it holds
$\Pi_1 \models \Gamma$ if and only if $\Pi_2 \models \Gamma$.
We denote var-equivalence between $\Pi_1$ and $\Pi_2$
by $\Pi_1 \qequiv{V} \Pi_2$.

\end{definition}

If $V$ is the set of all variables,
$\equiv$ and $\qequiv{V}$ coincide. On the other
hand, if $V$ is only composed of a subset of the
variables, these two kinds of equivalence are
different. In particular, while checking equivalence
is \conp-complete, checking var-equivalence is
\P{2}-complete \cite{lang-etal-03}. As a result,
checking var-redundancy is expected to be different
from redundancy, and to be harder. The following
equivalent condition of var-equivalence simplifies
the subsequent proofs.

\begin{theorem}

$\Pi_1 \qequiv{V} \Pi_2$ holds if and only if, for any
cube $\delta$ over $V$ (\ie, a non-tautological
clause containing all variables over $V$),
it holds $\Pi_1 \models \delta$ if and only if
$\Pi_2 \models \delta$.

\end{theorem}

\proof Follows from the fact that any formula over
variables $V$ can be expressed as a conjunction of
cubes over $V$.~\qed

This theorem simply tells us that equivalence can be
checked by looking at the cubes over $V$, rather than
checking all possible formulae. This theorem also
implies that all formulae that are var-equivalent are
also var-equivalent to some formulae that only
contain variables of $V$: one such formula is the
disjunction of all cubes over $V$ that are implied.
This formula is called the forgetting of the variables
that are not in $V$ \cite{lang-etal-03}.

Since cubes correspond to models, a
similar property based on partial models holds. To
this aim, we have however to give a special definition
of model satisfaction.

\begin{definition}[Var-models]

A model $\omega_V$ over variables $V$ is a var-model
of $\Pi$ if and only if the set of literals implied
by $\omega_V$ is consistent with $\Pi$. We denote this
fact as $\omega_V \stackrel{V}{\models} \Pi$.

\end{definition}

In other words, $\omega_V$ is a var-models of $\Pi$
if and only if there exists another model $\omega'$ 
over the set of variables not in $V$ such that
$\omega_V \omega' \models \Pi$. Using this definition
of models, we can give a semantical characterization
of var-equivalence.

\begin{theorem}

$\Pi_1 \qequiv{V} \Pi_2$ holds if and only if,
for any model $\omega_V$ over $V$ it
holds $\omega_V \stackrel{V}{\models} \Pi_1$
if and only if $\omega_V \stackrel{V}{\models} \Pi_2$.

\end{theorem}

The definition of var-redundancy differs from
that of redundancy only because logical equivalence
is replaced by var-equivalence.

\begin{definition}[Var-Redundancy of a Clause]

A clause $\gamma$ is var-redundant in $\Pi$
\wrt\  variables $V$ if and only if
$\Pi \backslash \{\gamma\} \qequiv{V} \Pi$.

\end{definition}

The fact that var-redundancy is different from
redundancy can be seen from the following formula
using $V=\{x\}$:

\[
\Pi = \{ x, y \}
\]

$\Pi$ is logically irredundant. However, the clause $y$
is var-redundant in $\Pi$ \wrt\
$V=\{x\}$: if queries are restricted to formulae
built on the variable $x$ only, then the clause $y$
is not needed. Note that var-redundancy does not
depend only on the variables a clause contains:
the clause $\gamma = \neg y$ is not var-redundant
in the set $\Pi=\{x \vee y, \neg y\}$ \wrt\
$V=\{x\}$ even if it does not mention any variable in $V$.

\begin{definition}[Var-Redundancy of a Set]

A set of clauses is var-redundant if and only
if it contains a clause that is var-redundant
in it.

\end{definition}

Since entailment is monotonic, var-irredundancy
of all clauses of $\Pi$ is the same as the non-equivalence
of $\Pi$ with one of its proper subsets. The problem is
therefore not harder than the problem of equivalence, as
a linear number of equivalence checks that can be done
in parallel are as hard as a single one.

Since var-equivalence is harder than logical equivalence
(\P{2}-complete \cite{lang-etal-03} vs.\  \conp-complete),
we expect var-redundancy to be harder than logical redundancy.
However, it is also easy to prove that redundancy is in the same
class of the corresponding equivalence problem, as it
amounts to solve a number of equivalence problems that can
be done in parallel. Proving that var-redundancy is hard for
the same class, instead, is slightly more difficult.
The following property is useful.

\begin{lemma}

A clause $\gamma$ is var-redundant in $\Pi$ \wrt\
$V$ if and only if any var-model of
$\Pi \backslash \{\gamma\}$ over $V$ is a also a
var-model of $\Pi$.

\end{lemma}

If a clause $\gamma$ only contains variables of $V$,
checking redundancy is relatively easy, as it amounts
to checking whether $\gamma$ is logically implied by
the other clauses. As a result, the \P{2}-hardness of
the problem of redundancy of a single clause can only
be proved if the clause contains some literals not in
$V$.

\begin{theorem}

Checking whether a clause
is var-redundant \wrt\  $V$ in a set
is \P{2}-complete.

\end{theorem}

\proof Membership follows from the fact that checking
var-equivalence is in \P{2}. Hardness is proved by
showing that $(\neg a \vee \Sigma) \cup \{a\}$
is var-equivalent \wrt\  $X$ to $(\neg a \vee \Sigma)$
if and only if $\forall X \exists Y . \Sigma$, where
$(\neg a \vee \Sigma)$ denotes the set
$\{\neg a \vee \gamma ~|~ \gamma \in \Sigma\}$.

\begin{description}

\item[Assume $\forall X \exists Y ~.~ \Sigma$.]
This assumption can be rephrased as: all partial
models over $X$ can be extended to form a model of
$\Sigma$. All models over $X$
can then be extended to form a model that satisfies
both $(\neg a \vee \Sigma) \cup \{a\}$ and
$(\neg a \vee \Sigma)$ by simply adding the
evaluation of $a$ to $\true$, that is, all var-models
of $\neg a \vee \Sigma$ are var-models of
$(\neg a \vee \Sigma) \cup \{a\}$.

\item[Assume $\exists X \forall Y ~.~ \neg \Sigma$.]
We prove that $(\neg a \vee \Sigma) \cup \{a\}$ and
$(\neg a \vee \Sigma)$ are not equivalent.
Let $\omega_X$ be the model over $X$ such that
$\Sigma$ is false regardless of the value of $Y$.
We show that $\omega_X$ is a var-model of
$(\neg a \vee \Sigma)$, but not of the other
formula. Extending the model $\omega_X$ with the model
that sets $a$ to $\false$ and $Y$ to any value, we
obtain a model of $(\neg a \vee \Sigma)$ simply
because all clauses in this set contains $\neg a$.

Let us now prove that $\omega_X$ is not a var-model
of $(\neg a \vee \Sigma) \cup \{a\}$, \ie, it
cannot be extended
to form a model of $(\neg a \vee \Sigma) \cup \{a\}$.
By the contrary, let $\omega_Y$ be the partial model
of $Y$ such that $\omega_X \omega_Y \omega_a$ satisfies
this formula. Since the formula contains $a$, the
model $\omega_a$ must set $a$ to true. As a result,
the formula can be reduced to $\Sigma$. This implies
that there exists $\omega_Y$ that extends $\omega_X$
to form a model of $\Sigma$, contradicting the
assumption.

\end{description}~\qed

The following theorem shows the complexity of
var-redundancy of a set of clauses. This problem
has the same complexity of var-redundancy
of a single clause, as in the case of logical
redundancy.

\begin{theorem}

Checking var-redundancy of a set of clauses
is \P{2}-complete.

\end{theorem}

\proof Membership follows from the fact that a set
is var-redundant if and only if
$\Pi \backslash \{\gamma\} \qequiv{V} \Pi$ holds
for some clause $\gamma \in \Pi$. These queries can
be done in parallel. Therefore, the problem is in
the same class of the single test, which is in \P{2}.

Hardness is proved by showing that
$\forall X \exists Y ~.~ \Sigma$ holds if and only if 
the following set $\Pi$ is var-redundant \wrt\
$V = X \cup B \cup C$, where
$\Sigma=\{\gamma_1,\ldots,\gamma_m\}$. We assume,
without loss of generality, that $\Sigma$ contains
at least two clauses, and it does not contain any
tautological clause.

\begin{eqnarray*}
\Pi &=& 
\{\pi\} \cup
\bigcup_{i=1,\ldots,m} \Pi_i
\\
&& \mbox{where}
\\
\pi &=&
\neg c_1 \vee \cdots \vee \neg c_m \vee a
\\
\Pi_i &=&
\{c_i \rightarrow \neg a \vee \gamma_i\} \cup
\{\neg a \rightarrow b_i\} \cup
\{l_j \rightarrow b_i ~|~ l_j \in \gamma_i\}
\end{eqnarray*}

Each set $\Pi_i$ entails the clause $c_i \rightarrow b_i$
This is indeed the result of resolving all clauses of
$\Pi_i$ together. As a result, $c_i \rightarrow b_i$ is
a consequence of $\Pi$. Being composed of variables of $V$
only, this clause must also be entailed by any
var-equivalent formula.

\begin{description}

\item[All clauses of $\Pi_i$ are irredundant in $\Pi$.]
This is proved by showing that, removing one clause of
$\Pi_i$ from $\Pi$, a new var-model is created. Since
$\Pi$ entails $c_i \rightarrow b_i$, the following partial
model cannot be extended to form a var-model of $\Pi$,
as it evaluates $c_i$ to $\true$ but $b_i$ to $\false$.

\[
\omega_{BC} = \{c_i,\neg b_i\} \cup
\{\neg c_j, b_j ~|~ j \in \{1,\ldots,m\} \backslash i\}
\]

We prove that, removing a clause $\delta \in \Pi_i$,
this model can be extended to form a model, that is,
$\omega_{BC}$ can be extended to form a model of
$\Pi \backslash \{\delta\}$. Since $m \geq 2$ by
assumption, $\omega_{BC}$ evaluates a variable
$c_j$ to $\false$, and the clause $\pi$ is therefore
satisfied. The model $\omega_{BC}$ also satisfies all
sets $\Pi_j$ with $j \not= i$. We therefore
only have to prove that, removing a clause
from $\Pi_i$, the model $\omega_{BC}$ can be
extended to form a model of the other clauses of $\Pi_i$.

\begin{description}

\item[$c_i \rightarrow \neg a \vee \gamma_i$\rm :]
the model $\omega$ with $\omega(a)=\true$ and
$\omega(l_j)=\false$ for any $l_j \in \gamma_i$
is such that
$\omega_{BC}\omega \models \Pi_i \backslash
\{c_i \rightarrow \neg a \vee \gamma_i\}$;

\item[$\neg a \rightarrow b_i$\rm :] the model
$\omega$ with $\omega(a)=\false$ and
$\omega(l_j)=\false$ for any $l_j \in \gamma_i$
is such that
$\omega_{BC}\omega \models \Pi_i \backslash
\{\neg a \rightarrow b_i\}$;

\item[$l_j \rightarrow b_i$\rm :] we use the
model $\omega$ with $\omega(a)=\true$,
$\omega(l_j)=\true$, and $\omega(l_k)=\false$
for any $l_k \in \gamma_i$ with $k \not= j$.
Indeed,
$\omega_{BC}\omega \models \Pi_i \backslash
\{l_j \rightarrow b_i\}$.

\end{description}

As a result, all clauses of $\Pi_i$ are irredundant
in $\Pi$. In other words, $\Pi$ is redundant if and
only if $\pi$ is redundant in $\Pi$.

\item[Assume $\exists X \forall Y ~.~ \neg \Sigma$.]
We prove that $a$ is irredundant. This is proved by
showing that $\Pi \backslash \{\pi\}$ has a var-model
that $\Pi$ has not. This var-model is $\omega_X\omega_{BC}$,
where $\omega_X$ is the value of $X$ that makes
$\Sigma$ falsified, while $\omega_{BC}$ evaluates
all variables in $B$ and $C$ to $\true$. This model
can indeed by extended to form a model of
$\Pi \backslash \{\pi\}$ by simply setting $a$ to
false. On the other hand, assume that there exists
$\omega_Y\omega_a$ such that
$\omega_X\omega_{BC}\omega_Y\omega_a \models \Pi$.
Since $\pi \in \Pi$, and $\omega_{BC}$ evaluates
all $c_i$'s to $\true$, we have $\omega_a(a)=\true$.
As a result, all clauses $c_i \rightarrow \neg a \vee \gamma_i$
can be simplified to $\gamma_i$. We can therefore
conclude that $\omega_X\omega_Y \models \Sigma$,
contrarily to the assumption.

\item[Assume that $\pi$ is irredundant in $\Pi$.]
We show that there exists $\omega_X$ that falsifies $\Sigma$
regardless of the value of $\omega_Y$. By assumption, there
is a var-model of $\Pi \backslash \{\pi\}$ that is not
a var-model of $\Pi$. Let $\omega_X\omega_{BC}$ be
such a var-model.

If $\omega_{BC}(c_i)=\false$ for some $i$, then
$\omega_{BC} \models \pi$. Since $\omega_{BC}$ is 
a var-model of $\Pi \backslash \{\pi\}$, there exists
$\omega$ such that
$\omega_{BC}\omega \models \Pi \backslash \{\pi\}$. Since
$\omega_{BC} \models \pi$, we also have that
$\omega_{BC}\omega \models \Pi$, contradicting the
assumption that $\omega_{BC}$ is not a var-model of
$\Pi$. As a result $\omega_{BC}(c_i)=\true$
for all indexes $i$. Since $c_i \rightarrow b_i$ is
entailed by $\Pi_i$ and, therefore, by
$\Pi \backslash \{\pi\}$, we can conclude that
$\omega_{BC}$ evaluates to $\true$ all variables
of $B \cup C$.

As a result, all formulae $\neg a \rightarrow b_i$
and $l_j \rightarrow b_i$ are satisfied by
$\omega_{BC}$.
Moreover, $c_i \rightarrow \neg a \vee \gamma_i$
simplifies to $\neg a \vee \gamma_i$, and $\pi$
simplifies to $a$. Since $\omega_X\omega_{BC}$
is not a var-model of $\Pi$, then $\omega_X$ is not
a var-model of $\{\neg a \vee \gamma_i\} \cup \{a\}$,
which is equivalent to $\Sigma$. In other words,
$\omega_X$ cannot be extended to form a model of
$\Sigma$.

\end{description}~\qed

\subsection{Conditional Equivalence}

Conditional equivalence (or strong equivalence) of
two formulae holds whenever the two formulae are
equivalent and remain so regardless of updates.
This definition depends on how revisions of knowledge
bases are done. If the semantics of revision is
syntax-independent \cite{dala-88},
then conditional and logical equivalence coincide.
On the other hand, objections to the principle of
the irrelevance of syntax have been raised
\cite{fuhr-91,nebe-91,bess-etal-99,wass-01}, and some
revision semantics that depend on the syntax exist. They are
mainly motivated by the fact that the syntactic form in which
a formula is expressed tells more than its set of models.
In this section, we only consider the basic definition
of revision by Fagin, Ullman, and Vardi
\cite{fagi-ullm-vard-83} and by Ginsberg \cite{gins-86}.

\begin{definition}

$Max(\Pi,\Gamma)$ is the set of the maximal subsets
of $\Pi$ that are consistent with $\Gamma$. Revision
of $\Pi$ with $\Gamma$ is defined as follows:

\[
\Pi * \Gamma = \bigvee Max(\Pi,\Gamma)
\]

\end{definition}

We now give an equivalent characterization of this form of
revision. Given a set of clauses $\Pi$, let $S_\Pi(\omega)$
be the set of clauses of $\Pi$ that are satisfied
by the model $\omega$.

\begin{definition}[Satisfied Subset]

The subset of $\Pi$ satisfied by $\omega$ is:

\[
S_\Pi(\omega) = \{\gamma \in \Pi ~|~ \omega \models \gamma\}
\]

\end{definition}

The result of revision can be characterize in terms
of the set of models of the result.

\begin{lemma}
\label{revision-maximal}

The models of $\Pi * \Gamma$ are exactly the
models $\omega$ of $\Gamma$ whose set $S_\Pi(\omega)$
is maximal with respect to set containment.

\end{lemma}

\proof By definition, only models of $\Gamma$ have
to be taken into account. The set $S_\Pi(\omega)$ is the
set of formulae of $\Pi$ that are satisfied
by $\omega$. Since $\omega$ is a model of $\Gamma$, we have
that $S_\Pi(\omega) \cup \Gamma$ is consistent. As
a result, the only case in which $\omega$ is not a model
of the revision is when this set is not one of those
maximally consistent with $\Gamma$, that is, there
exists a maximal $\Pi'$ such that $\Pi' \cup \Gamma$
is consistent, and $S_\Pi(\omega) \subset \Pi'$. Since
$\Pi' \cup \Gamma$ is consistent, it has models:
let $\omega'$ be a model of $\Pi' \cup \Gamma$. Since
all clauses of $\Pi'$ satisfy $\omega'$, and $\Pi'$
is maximally consistent, we have
$\Pi' = S_\Pi(\omega')$. This proves that $\omega$ is
not a model of $\Pi * \Gamma$ if and only if there
exists $\omega'$ with $S_\Pi(\omega) \subset S_\Pi(\omega')$.~\qed

We can now formally give the definition of conditional
redundancy of a clause, and of a set of clauses.

\begin{definition}[Conditional Redundancy of a Clause]

A clause $\gamma$ is conditionally redundant
in $\Pi$ if and only if $\Pi$ is conditionally
equivalent to $\Pi \backslash \{\gamma\}$.

\end{definition}

The definition of conditional redundancy of a
whole set can be defined in two different ways:
first, a set is conditionally redundant if it
contains a redundant clause; second, a set
is conditionally redundant if it is equivalent
to one of its proper subsets. We use the first
definition.

\begin{definition}[Conditional Redundancy of a set of Clauses]

A set of clauses $\Pi$ is conditionally redundant
if and only if it contains a redundant clause.

\end{definition}

This definition is not equivalent to the other
one. For example, the set
$\Pi = \{a \vee b, a \vee \neg b,
a \vee c, a \vee \neg c\}$ does not contain any
conditionally redundant clause, but is conditionally
equivalent to its subset $\Pi'=\{a \vee b, a \vee \neg b\}$.
This difference is caused by the fact that revision
is a non-monotonic operator: most, but not all,
non-monotonic logics show this phenomena \cite{libe-redu-3}.

Lemma~\ref{revision-maximal} tells that a clause is redundant
if and only if its removal modifies the ordering
on models defined by $\subseteq$ on $S_\Pi(.)$. On
the other hand, it may be that two models are incomparable
before the removal of a clause and equal afterwards.
As a result, the difference of the orderings caused
is only a necessary condition to equivalence, not
a sufficient one.

\begin{lemma}
\label{conditional-ordering-necessary}

If $\gamma$ is irredundant in $\Pi$, then there exists two
models $\omega$ and $\omega'$ such that the containment
relation between $S_\Pi(\omega)$ and $S_\Pi(\omega')$
is different from that between
$S_{\Pi \backslash \{\gamma\}}(\omega)$ and
$S_{\Pi \backslash \{\gamma\}}(\omega')$.

\end{lemma}

\proof Trivial consequence of Lemma~\ref{revision-maximal}:
if the ordering is the same, then all revision results
are the same.~\qed

This condition is however not a sufficient one, in
general: indeed, it may be that the ordering is different
only because two sets that are incomparable becomes
equal. If this is the case, the result of revision is
always the same. On the other hand, such a case is not
possible if the two formulae only differ for one clause.

\begin{lemma}
\label{conditional-ordering}

$\gamma$ is irredundant in $\Pi$ if and only if
there exists two
models $\omega$ and $\omega'$ such that the containment
relation between $S_\Pi(\omega)$ and $S_\Pi(\omega')$
is different from that between
$S_{\Pi \backslash \{\gamma\}}(\omega)$ and
$S_{\Pi \backslash \{\gamma\}}(\omega')$.

\end{lemma}

\proof The ``only if'' part is
Lemma~\ref{conditional-ordering-necessary}. We only
have to prove that, if the containment
relation between $S_\Pi(\omega)$ and $S_\Pi(\omega')$
is different from that between
$S_{\Pi \backslash \{\gamma\}}(\omega)$ and
$S_{\Pi \backslash \{\gamma\}}(\omega')$, then
$\gamma$ is irredundant.

If both $\omega$ and $\omega'$ satisfy $\gamma$, then
its removal does not change the relationship between
$S_\Pi(\omega)$ and $S_\Pi(\omega')$, as $\gamma$ is
removed from both. On the other hand, if none of these
models satisfy $\gamma$, then the sets $S_\Pi(\omega)$
and $S_\Pi(\omega')$ are not modified at all.

The only remaining case is therefore that one of these
two models satisfy $\gamma$ while the other does not.
Without loss of generality, assume that $\omega$
satisfies $\gamma$ while $\omega'$ does not. As an
immediate result, we have that
$S_\Pi(\omega) \not\subseteq S_\Pi(\omega')$, since the
first set contains a clause the other one does not.
Moreover, we have that

\begin{eqnarray*}
S_{\Pi \backslash \{\gamma\}}(\omega)
&=&
S_\Pi(\omega) \backslash \{\gamma\}
\\
S_{\Pi \backslash \{\gamma\}}(\omega') &=& S_\Pi(\omega')
\end{eqnarray*}

In other words, the only effect of removing $\gamma$
is to remove $\gamma$ from the set of clauses that
are satisfied by $\omega$, while the clauses satisfied
by $\omega'$ are the same.

We prove that the inverse containment is not modified
by the removal of $\gamma$. Formally, we prove that
$S_\Pi(\omega') \subseteq S_\Pi(\omega)$ if and only
if $S_{\Pi \backslash \{\gamma\}}(\omega') \subseteq
S_{\Pi \backslash \{\gamma\}}(\omega)$.

\begin{enumerate}

\item If $S_\Pi(\omega') \subseteq S_\Pi(\omega)$ holds,
using the equations above we have that
$S_{\Pi \backslash \{\gamma\}}(\omega') \subseteq
S_{\Pi \backslash \{\gamma\}}(\omega) \cup \{\gamma\}$.
Since $\gamma \not\in S_{\Pi \backslash \{\gamma\}}(\omega')$,
this is equivalent to
$S_{\Pi \backslash \{\gamma\}}(\omega') \subseteq
S_{\Pi \backslash \{\gamma\}}(\omega)$.

\item If
$S_{\Pi \backslash \{\gamma\}}(\omega') \subseteq
S_{\Pi \backslash \{\gamma\}}(\omega)$, by using the
equations above, we have that
$S_\Pi(\omega') \subseteq S_\Pi(\omega) \backslash \{\gamma\}$,
which implies that
$S_\Pi(\omega') \subseteq S_\Pi(\omega)$.

\end{enumerate}

We can therefore conclude that the only possible
change of relationship between the clauses that
are satisfied by $\omega$ and those satisfied by
$\omega'$ is that 
$S_\Pi(\omega) \not\subseteq S_\Pi(\omega')$
but $S_{\Pi \backslash \{\gamma\}}(\omega) \subseteq
S_{\Pi \backslash \{\gamma\}}(\omega')$, while
the set containment in the other direction is preserved.

Let $\Gamma$ be the formula that has $\omega$ and $\omega'$
has its only two models. We prove that the revision
by $\Gamma$ is affected by the presence of $\gamma$. Formally,
we show that $\Pi * \Gamma$ is different from
$\Pi \backslash \{\gamma\} * \Gamma$. We have already shown
that $S_\Pi(\omega) \not\subseteq S_\Pi(\omega')$,
and that the inverse containment is not changed by the
removal of $\gamma$. Since the containment relation
changes by assumption, we also have that
$S_{\Pi \backslash \{\gamma\}}(\omega) \subseteq
S_{\Pi \backslash \{\gamma\}}(\omega')$. Two cases are
possible: either $S_\Pi(\omega') \subseteq S_\Pi(\omega)$
or not. Let us consider each case separately.

\begin{itemize}

\item
If $S_\Pi(\omega') \subseteq S_\Pi(\omega)$, since
$S_\Pi(\omega) \not\subseteq S_\Pi(\omega')$,
we have $S_\Pi(\omega') \subset S_\Pi(\omega)$. As
a result $\Pi * \Gamma$ has $\omega$ as its only model.
On the other hand,
$S_{\Pi \backslash \{\gamma\}}(\omega) \subseteq
S_{\Pi \backslash \{\gamma\}}(\omega')$. As a result,
$\omega$ and $\omega'$ are evaluated in the same
way by $S_{\Pi \backslash \{\gamma\}}(.)$.
As a result, $\Pi \backslash \{\gamma\} * \Gamma$ has both
of them as models.

\item
If $S_\Pi(\omega') \not\subseteq S_\Pi(\omega)$,
since $S_\Pi(\omega) \not\subseteq S_\Pi(\omega')$,
we have that $\omega$ and $\omega'$ are incomparable
in $\Pi$. As a result, $\Pi * \Gamma$ has both of them
as models.

On the other hand, we have that
$S_{\Pi \backslash \{\gamma\}}(\omega) \subseteq
S_{\Pi \backslash \{\gamma\}}(\omega')$, and therefore
$S_{\Pi \backslash \{\gamma\}}(\omega) \subset
S_{\Pi \backslash \{\gamma\}}(\omega')$. As a result,
$\omega'$ is strictly preferred over $\omega$ in
$\Pi \backslash \{\gamma\}$. As a result,
$\Pi \backslash \{\gamma\} * \Gamma$ has $\omega'$ as its
only model.

\end{itemize}

We have therefore proved the following: if the removal
of $\gamma$ changes the relationship between two models
$\omega$ and $\omega'$, then the only possible change
is that $S_\Pi(\omega) \not\subseteq S_\Pi(\omega')$ and
$S_{\Pi \backslash \{\gamma\}}(\omega) \subseteq
S_{\Pi \backslash \{\gamma\}}(\omega')$, while the
inverse containment relationship is not changed. We have
then proved that such a change leads to different
results when $\Pi$ and $\Pi \backslash \{\gamma\}$ are
both revised by the same formula $\Gamma$. As a result,
$\gamma$ is irredundant.~\qed

This lemma proves that the irredundancy of a clause
is related to the modification of the set containment
of the sets of clauses that are satisfied by the models.
On the other hand, this condition is only about the
redundancy of a single clause. If we allow removing
two clauses, the ordering can be modified while
conditional equivalence is preserved.

\begin{theorem}
\label{cond-one-more-not}

The following two sets of clauses are conditionally
equivalent, but the ordering they induce are different,
and all clauses of $\Pi$ are irredundant.

\begin{eqnarray*}
\Pi &=&
\{a \vee b, a \vee \neg b,
a \vee c, a \vee \neg c\} \\
\Pi' &=& \{a \vee b, a \vee \neg b\}
\end{eqnarray*}

\end{theorem}

\proof We prove that $\Pi$ does not contain any
redundant clause. Its symmetry allows proving it
for a single clause only. Let us therefore show
that $a \vee b$ is irredundant. Consider the
following revising formula $\Gamma=\neg a$. The maximal
subsets of $\Pi$ that are consistent with $\Gamma$
are composed of exactly one clause between
$a \vee b$ and $a \vee \neg b$, and one clause
between $a \vee c$ and $a \vee \neg c$. As a
result, $\Pi * \Gamma = \neg a$.

Let us now consider $\Pi \backslash \{a \vee b\} * \Gamma$.
The maximal subsets of $\Pi \backslash \{a \vee b\}$
that are consistent with $\neg a$ contains the
clause $a \vee \neg b$, and one clause between
$a \vee c$ and $a \vee \neg c$. As a result, all
maximal subset contains $a \vee \neg b$, which is
therefore in $\Pi \backslash \{a \vee b\} * \Gamma$. We
can therefore conclude that
$\Pi \backslash \{a \vee b\} * \Gamma = \neg a \wedge \neg b$.
Since this is different from $\Pi * \Gamma$, the clause
$a \vee b$ is irredundant in $\Pi$.

We now prove that $\Pi'$ is conditionally equivalent
to $\Pi$. Let $\omega$ and $\omega'$ be two models.
If they both satisfy $c$ or they both satisfy
$\neg c$, the set of satisfied clauses are modified
in the same way. On the other hand, if one of them
implies $c$ and the other one implies $\neg c$, then
they are incomparable in $\Pi$, but equal in $\Pi'$.
The only difference is therefore that some pairs of
models are incomparable in $\Pi$ but equal in $\Pi'$.
As a result, the maximal ones are always the same.

While $\Pi$ and $\Pi'$ are conditionally equivalent,
there exist two models that are compared
differently in $\Pi$ and $\Pi'$.
Let $\omega$ and $\omega'$ be the models such
that $\omega \models \neg a \wedge \neg b \wedge \neg c$
and $\omega' \models \neg a \wedge \neg b \wedge c$.
The sets $S_\Pi(\omega)$ and $S_\Pi(\omega')$
are not comparable: the first contains $a \vee \neg c$
but not $a \vee c$, while the second contains the second
one but not the first. As a result, the ordering is
changed.~\qed

The following lemma makes the statement of
Lemma~\ref{conditional-ordering} more precise: not
only there is a pair of models whose ordering is
modified: this ordering is modified in a very
specific way.

\begin{lemma}
\label{conditional-onemore}

$\gamma$ is conditionally irredundant in $\Pi$ if and
only if there exists two models $\omega$ and $\omega'$
such that:

\[
S_\Pi(\omega) \backslash S_\Pi(\omega') = \{\gamma\}
\]

\end{lemma}

\proof If two such model exists, then we have that
$S_\Pi(\omega) \not\subseteq S_\Pi(\omega')$, since
$S_\Pi(\omega)$ contains a clause that is not in
$ S_\Pi(\omega')$; on the other hand, since $\gamma$
is the only clause that is in  $S_\Pi(\omega)$ but not
in $S_\Pi(\omega')$, removing it from both sets leads to
$S_{\Pi \backslash \{\gamma\}}(\omega) \subseteq
S_{\Pi \backslash \{\gamma\}}(\omega')$. This result
tells us that the removal of $\gamma$ modifies the
relationship between the set of clauses that are
satisfied by $\omega$ and by $\omega'$. By
Lemma~\ref{conditional-ordering}, this implies that
$\gamma$ is irredundant.

Let us assume that $\gamma$ is irredundant. By
Lemma~\ref{conditional-ordering-necessary},
there are two models $\omega$ and $\omega'$ such that
the containment relation between $S_\Pi(\omega)$
and $S_\Pi(\omega')$ is affected by the presence of
$\gamma$ in $\Pi$.
If $\omega$ and $\omega'$ evaluate
$\gamma$ in the same way (\ie, either both or none
of them satisfy it), then removing $\gamma$ modifies their
sets $S_\Pi(.)$ in the same way (either $\gamma$ is
removed from both, or it is not in either already.)

As a result, either $\omega$ or $\omega'$ satisfy
$\gamma$, but not both. Without loss of generality,
we can assume that $\omega$ is the model that
satisfy $\gamma$. As a result, we have that
$\gamma \in S_\Pi(\omega) \backslash S_\Pi(\omega')$.
We therefore only have to prove that no other
clause is in this difference. On the converse, assume
that $S_\Pi(\omega) \backslash S_\Pi(\omega')$
contains another clause $\gamma'$, that is
$\{\gamma, \gamma'\} \subseteq S_\Pi(\omega) \backslash S_\Pi(\omega')$.
The only effect of removing $\gamma$ is that $\gamma$
disappears from the set of clauses satisfied by
$\omega$; on the other hand, $\gamma'$ is still
there. As a result, the relationship between the
set of satisfied clauses remains the same. Formally,
two cases are possible: either $\omega$ satisfies
all clauses that are satisfied by $\omega'$, or
$\omega'$ satisfies some clauses more. In the first
case, the removal of $\gamma$ does not change the
relationship because $\omega$ still satisfies all
clauses of $\omega'$ and $\gamma'$. In the second
case, the sets of satisfied clauses are still
incomparable, as $\omega'$ satisfies the same clauses,
while $\omega$ satisfies $\gamma'$.~\qed

We have now all technical tools to prove the
complexity of checking redundancy of a single
clause in a set.

\begin{theorem}
\label{theorem-conditional-one}

Checking whether $\gamma$ is conditionally redundant
in $\Pi$ is \conp-complete.

\end{theorem}

\proof Membership: a clause is redundant if and only
if there exists two models such that their ordering
is affected by the presence of the clause.

Hardness: the set of clauses $\Pi$ is satisfiable if
and only if the clause $a$ is conditionally redundant
in $\Sigma=(a \vee \Pi) \cup \{a\}$, where
$a \vee \Pi$ is a shorthand for
$\{a \vee \gamma ~|~ \gamma \in \Pi\}$. We divide the proof
in two parts: first, we consider the case in which $\Pi$
is satisfiable, and prove that $a$ is irredundant;
second, we show that the irredundancy of $a$ implies the
satisfiability of $\Pi$.

If $\Pi$ is satisfiable, it has a model $\omega_X$.
We show that $\gamma$ is irredundant in $\Sigma$ by
considering two models: the first one is $\omega$,
which is obtained by adding the evaluation $a=\false$ to
$\omega_X$; the second one is $\omega_T$, the model
that sets all variables to $\true$. The first model
satisfies all clauses but $a$; the second model
satisfies all clauses. As a result, we have that
$a$ is the only clause satisfied by $\omega_T$ that
is not satisfied by $\omega$, that is:
$S_\Sigma(\omega_T) \backslash S_\Sigma(\omega) = \{a\}$.
By Lemma~\ref{conditional-onemore}, this implies that
$a$ is irredundant.

Let us now assume that $a$ is irredundant. By
Lemma~\ref{conditional-onemore},
$S_\Sigma(\omega) \backslash S_\Sigma(\omega')$
is equal to $\{a\}$ for some pair of models
$\omega$ and $\omega'$. This condition implies that
$\omega$ satisfies $a$ while $\omega'$ does not.
Since $\omega$ satisfies $a$ we have that
$S_\Sigma(\omega)=\Sigma$ since all clauses of $\Sigma$
contains $a$. As a result,
$S_\Sigma(\omega') = \Sigma \backslash \{a\}$, that
is, $\omega'$ satisfies all clauses of $a \vee \Pi$.
Since $\omega'(a)=\false$, the model $\omega'$ satisfies
all clauses of $\Pi$.~\qed

We now prove that checking whether a formula
contains a redundant clause is \conp-complete
as well. Note that the redundancy of a formula
is defined as the presence of a redundant clause
in the set, and not as the property of being
equivalent to a proper subset. These two definitions
are not equivalent, as shown by
Theorem~\ref{cond-one-more-not}.

\begin{theorem}
\label{theorem-conditional-all}

Checking redundancy (\ie, presence of a redundant
clause) of a set of clauses is \conp-complete.

\end{theorem}

\proof Membership is proved as usual: we have to
check the redundancy of some clause; these tests
can be done in parallel, and therefore the whole
problem is in \conp.

Hardness is proved as follows: we prove that the clause
$a$ is redundant in the following set $\Sigma$ if and
only if $\Pi$ is unsatisfiable:

\[
\Sigma =
\{\neg c_i \vee a \vee \gamma_i ~|~ \gamma_i \in \Pi \} \cup
\{c_i \vee a ~|~ \gamma_i \in \Pi \} \cup
\{a\}
\]

We indeed prove the following: first, all clauses but $a$
are irredundant. Second, that $a$ is redundant if and only
if $\Pi$ is satisfiable.

\begin{description}

\item[All clauses of $\Sigma \backslash \{a\}$ are irredundant.]
This is proved by showing, for each of them, a possible
revising formula $\Gamma$ such that $\Sigma * \Gamma$ is different
than $\Sigma \backslash \{\delta\} * \Gamma$ for each clause
$\delta$ of $\Sigma$ that is not $a$.

\begin{description}

\item[$\neg c_i \vee a \vee \gamma_i$\rm .] The formula
is $\Gamma=\neg a \wedge c_i \wedge \{\neg c_j ~|~ j \not= i \}$.
This formula satisfies $c_i \vee a$ and all clauses
$\neg c_j \vee a \vee \gamma_j$, and falsifies $a$ and
all clauses $c_j \vee a$. As a result, the only clause
that is not satisfied neither contradicted is
$\neg c_i \vee a \vee \gamma_i$. As a result, the result
of the revision entails $\gamma_i$ if and only if this clause
is present.

\item[$c_i \vee a$\rm .] We use the formula
$\Gamma = \neg a \wedge \{\neg c_j ~|~ j \not= i\} \wedge \gamma_i$.
This formula falsifies $a$, all clauses $c_j \vee a$
with $j \not= i$, and implies the clause
$\neg c_i \vee a \vee \gamma_i$ because of $\gamma_i$,
and all clauses $\neg c_i \vee a \vee \gamma_j$ for any
$j \not= i$ because $\Gamma \models \neg c_j$. As a result,
the only clause that is not falsifies nor entailed is
$c_i \vee a$. Its presence is needed to allow deriving
$\neg c_i$ from the revised theory.

\end{description}

\item[If $\Pi$ is satisfiable $a$ is irredundant.] This
is proved by showing a revising formula that makes $a$
needed for the entailment of some formulae. Namely,
since $\Pi$ is satisfiable it has a model $\omega$.
Let $\Gamma$ be defined as follows:

\[
\Gamma = \{c_i ~|~ \gamma_i \in \Pi \} \cup
\{x_i ~|~ \omega(x_i) = \true \} \cup
\{\neg x_i ~|~ \omega(x_i) = \false \}
\]

In words, we set all $c_i$'s to $\true$, and give to any
$x_i$ the sign that is in the model $\omega$. This formula
is clearly satisfiable. Moreover, it is almost complete,
since the only variable that is not forced to have a specific
value is $a$. Moreover, $\Gamma$ implies all clauses:
$\neg c_i \vee a \vee \gamma_i$ is implied because $\omega$
satisfies all clauses $\gamma_i$, while $c_i \vee a$ is
entailed because $\Gamma$ contains $c_i$. On the other hand, $a$
is not falsified nor it is entailed. As a result, the presence
of $a$ in the result of revision is related to its presence
in the original theory.

\item[If $a$ is irredundant, $\Pi$ is satisfiable.] This
is proved by using the characterization of irredundancy
provided by Lemma~\ref{conditional-onemore}: since $a$
is irredundant, there exists two models $\omega$ and
$\omega'$ such that:

\[
S_\Sigma(\omega) \backslash S_\Sigma(\omega') = \{a\}
\]

Therefore, $\omega \models a$ and $\omega' \not\models a$.
We can now proceed by using the following rules:

\begin{enumerate}

\item every clause but $a$ that is satisfied by $\omega$
is also satisfied by $\omega'$ (otherwise $a$ would not
be the only clause that is satisfied by $\omega$ but
not by $\omega'$);

\item every clause that is not satisfied by $\omega'$
is not satisfied by $\omega$ as well (same reason);

\item if a model satisfies some clauses, it also
satisfies all their consequences.

\end{enumerate}

This leads to the pictorial proof of
Figure~\ref{condred-sat}.

\begin{figure}[!htb]
\begin{center}
\setlength{\unitlength}{4144sp}%
\begingroup\makeatletter\ifx\SetFigFont\undefined%
\gdef\SetFigFont#1#2#3#4#5{%
  \reset@font\fontsize{#1}{#2pt}%
  \fontfamily{#3}\fontseries{#4}\fontshape{#5}%
  \selectfont}%
\fi\endgroup%
\begin{picture}(6268,6064)(-675,-6149)
\thinlines
\put(3691,-331){\vector( 0,-1){540}}
\put(1171,-331){\vector( 0,-1){990}}
\put(1801,-1681){\vector( 2,-1){1350}}
\put(721,-196){\line(-1, 0){720}}
\put(  1,-196){\line( 0,-1){4050}}
\put(  1,-4246){\vector( 1, 0){315}}
\put(1801,-4381){\vector( 2,-1){1170}}
\put(4771,-1006){\line( 1, 0){810}}
\put(5581,-1006){\line( 0,-1){5040}}
\put(5581,-6046){\vector(-1, 0){1395}}
\put(4771,-1006){\line( 0,-1){2520}}
\put(4771,-3526){\vector(-1, 0){585}}
\put(4366,-2536){\vector( 1, 0){405}}
\put(4231,-1006){\vector( 1, 0){540}}
\put(3691,-3661){\line( 0,-1){540}}
\put(3691,-4201){\vector( 1, 0){1890}}
\put(4501,-5146){\vector( 1, 0){1080}}
\put(1171,-241){\makebox(0,0)[b]{\smash{\SetFigFont{12}{14.4}{\familydefault}{\mddefault}{\updefault}$\omega \models a$}}}
\put(3691,-241){\makebox(0,0)[b]{\smash{\SetFigFont{12}{14.4}{\familydefault}{\mddefault}{\updefault}$\omega' \not\models a$}}}
\put(3691,-1051){\makebox(0,0)[b]{\smash{\SetFigFont{12}{14.4}{\familydefault}{\mddefault}{\updefault}$\omega' \models \neg a$}}}
\put(1171,-1591){\makebox(0,0)[b]{\smash{\SetFigFont{12}{14.4}{\familydefault}{\mddefault}{\updefault}$\omega \models c_i \vee a$}}}
\put(3691,-2581){\makebox(0,0)[b]{\smash{\SetFigFont{12}{14.4}{\familydefault}{\mddefault}{\updefault}$\omega' \models c_i \vee a$}}}
\put(3691,-3571){\makebox(0,0)[b]{\smash{\SetFigFont{12}{14.4}{\familydefault}{\mddefault}{\updefault}$\omega' \models c_i$}}}
\put(1171,-4291){\makebox(0,0)[b]{\smash{\SetFigFont{12}{14.4}{\familydefault}{\mddefault}{\updefault}$\omega \models \neg c_i \vee a \vee \gamma_i$}}}
\put(3691,-5191){\makebox(0,0)[b]{\smash{\SetFigFont{12}{14.4}{\familydefault}{\mddefault}{\updefault}$\omega' \models \neg c_i \vee a \vee \gamma_i$}}}
\put(3691,-6091){\makebox(0,0)[b]{\smash{\SetFigFont{12}{14.4}{\familydefault}{\mddefault}{\updefault}$\omega' \models \gamma_i$}}}
\end{picture}

\end{center}
\caption{If $a$ is irredundant then $\Pi$ is satisfiable}
\label{condred-sat}
\end{figure}
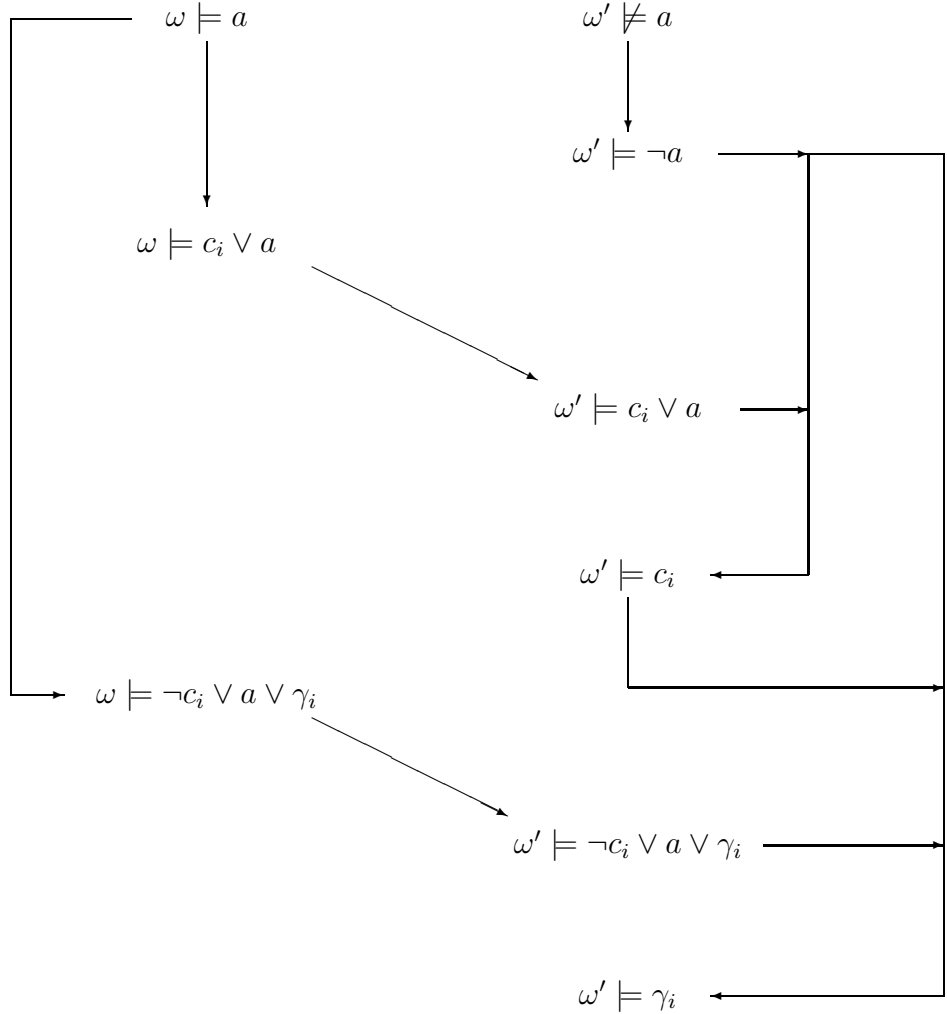

In words, the proofs proceeds as follows, using the
rules above and the fact that $\omega \models a$ and
$\omega' \not\models a$. The latter is equivalent to
$\omega' \models \neg a$. Since $\omega \models a$ we
have that $\omega \models c_i \vee a$. As a result,
the same clause is satisfied by $\omega'$, that is,
$\omega' \models c_i \vee a$. Since $\omega' \models \neg a$,
we can conclude that $\omega' \models c_i$ for all
indexes $i$.

Since $\omega \models a$, we also have that
$\omega \models \neg c_i \vee a \vee \gamma_i$. As
a result, $\omega'$ satisfies the same clause,
that is $\omega' \models \neg c_i \vee a \vee \gamma_i$.
But we have already proved that $\omega' \models \neg a$
and that $\omega' \models c_i$. As a result, we have
that $\omega' \models \gamma_i$ for all $i$. This
proves that $\omega'$ is a model of all clauses
$\gamma_i \in \Pi$. As a result, $\Pi$ is satisfiable.

\end{description}

We can therefore conclude that all clauses of $\Sigma$
but $a$ are irredundant, and that $a$ is redundant if
and only if $\Pi$ is unsatisfiable. As a result, $\Sigma$
is redundant if and only if $\Pi$ is unsatisfiable.~\qed

\let\subsectionnewpage=\relax

\section{Conclusions}

We have presented a study of the semantical and
computational properties of concepts related to the
redundancy of CNF propositional formulae. Namely, we have
considered the problem of checking whether a formula is
redundant and some problems related to removing
redundancy from it. The computational analysis has shown
that checking redundancy is \conp-complete. We have then
defined an \ies\  as an irredundant equivalent
subset of a formula, and studied some problems
related to \ies's: checking, size, uniqueness, and
membership of clauses to some or all \ies's.
All problems have been given an exact characterization
within the polynomial hierarchy, that is,
we have found classes these problems are complete for.
The problem of redundancy has also been studied for the
case of two alternative forms of equivalence based on
particular sets of possible queries.

Some problems are still open. Namely, irredundancy is only
one way of defining minimal representation of a formula,
but other ones exist. In the Horn case, several
different definitions of minimality have been used, both
by Meier~\cite{maie-80} and by Ausiello et
al.~\cite{ausi-etal-86}, including irredundancy and
number of occurrences of literals. In the
general (non-Horn) case, only the number of occurrences
of literals (and, in this paper, irredundancy) have been
considered. An open problem is whether the other notions
of minimality used in the Horn case make sense in the
general case as well.

Some other problems have not been considered in this paper,
and are analyzed in two other papers. In the first
one \cite{libe-redu-2}, the complexity of the problem of
redundancy has been analyzed for the case of Horn and
2CNF formulae. The analysis of 2CNF, in particular, has
shown a very interesting pattern: while the properties of
redundancy and irredundancy are different depending on
whether the formula implies some literals or not, a concept
of {\em acyclicity} makes often the difference between
tractability and intractability. In the other paper
\cite{libe-redu-3} some non-classical logics have been
considered: non-monotonic logics, multi-valued logics,
and logics for reasoning about actions. An interesting
issue of non-classical logics is that equivalence can be
defined in different ways, and that the irredundancy of
all parts of a knowledge base does not always imply the
irredundancy of the knowledge base.

\comment

Open problem:
  decide the existence of an equivalent subset of size <= k,
  where the size is number of occurrence of literals

\endcomment

\subsection*{Acknowledgments}

\iffalse

Many thanks to Marco Schaerf for his comments
on a previous version of this paper, and to the
anonymous reviewers for their very detailed
comments.
{\bf Thanks to: Chopra, Sattler, Furbach,
Van Harmelen, Farinas del Cerro?}

\else

[...]

\fi


\bibliographystyle{alpha}

\end{document}